\documentclass[10pt,twocolumn,letterpaper]{article}

\usepackage[pagenumbers]{cvpr} 

\usepackage[dvipsnames]{xcolor}

\usepackage{bm}
\usepackage{multirow}
\usepackage{arydshln}

\definecolor{cvprblue}{rgb}{0.21,0.49,0.74}
\usepackage[pagebackref,breaklinks,colorlinks,citecolor=cvprblue]{hyperref}

\def\paperID{*****} 
\def\confName{CVPR}
\def\confYear{2024}

\title{ShareCMP: Polarization-Aware RGB-P Semantic Segmentation}

\author{
Zhuoyan Liu \quad Bo Wang\thanks{Corresponding author.} \quad Lizhi Wang \quad Chenyu Mao \quad Ye Li \\
Harbin Engineering University \\
{\tt\small \{liuzhuoyan,wb,wanglizhi111,maochenyu,liye\}@hrbeu.edu.cn}
}

\begin{document}
\maketitle
\begin{abstract}
Multimodal semantic segmentation is developing rapidly, but the modality of RGB-\textbf{P}olarization remains underexplored. To delve into this problem, we construct a UPLight RGB-P segmentation benchmark with 12 typical underwater semantic classes. In this work, we design the ShareCMP, an RGB-P semantic segmentation framework with a shared dual-branch architecture (ShareCMP Encoder), which reduces the parameters and memory space by about 33.8\% compared to previous dual-branch models. It encompasses a Polarization Generate Attention (PGA) module designed to generate polarization modal images with richer polarization properties for the encoder. In addition, we introduce the Class Polarization-Aware Loss (CPALoss) with Class Polarization-Aware Auxiliary Head (CPAAHead) to improve the learning and understanding of the encoder for polarization modal information and to optimize the PGA module. With extensive experiments on a total of three RGB-P benchmarks, our ShareCMP achieves the best performance in mIoU with fewer parameters on the UPLight (92.45{\small (+0.32)}\%), ZJU (92.7{\small (+0.1)}\%), and MCubeS (50.99{\small (+1.51)}\%) datasets. And our ShareCMP (w/o PGA) achieves competitive or even higher performance on other RGB-X datasets compared to the corresponding state-of-the-art RGB-X methods. The code and datasets are available at \url{https://github.com/LEFTeyex/ShareCMP}.
\end{abstract}
\section{Introduction}
\label{sec:introduction}

Multimodal semantic segmentation provides the autonomous vehicle with scene understanding capability~\cite{mfnet,51} using perceptual modalities such as RGB-\textbf{D}epth, -\textbf{T}hermal, -\textbf{E}vent, and -\textbf{L}iDAR, and is also developing rapidly in recent years. In underwater scenes, Autonomous Underwater Vehicles (AUVs) also need this kind of scene understanding capability to perceive comprehensive environmental information and make correct decisions. Drawing inspiration from the use of polarized light by the mantis shrimp to perceive the underwater environment~\cite{52,53}, we investigate the modality of RGB-\textbf{P}olarization (RGB-P) into the perception systems of AUVs. The reliance of the mantis shrimp on polarization enables it to mitigate visual interference caused by light absorption, scattering, and low-light conditions. Similarly, our approach leverages this principle to enhance the perceptual capabilities of AUVs in complex underwater environments. However, despite the potential advantages of RGB-P for underwater applications, there is a notable lack of dedicated RGB-P semantic segmentation datasets for underwater environments. This hinders the development and benchmarking of effective methods, highlighting the need for further research in this area.

\begin{figure}[t]
\centering
\includegraphics[width=1.0\linewidth]{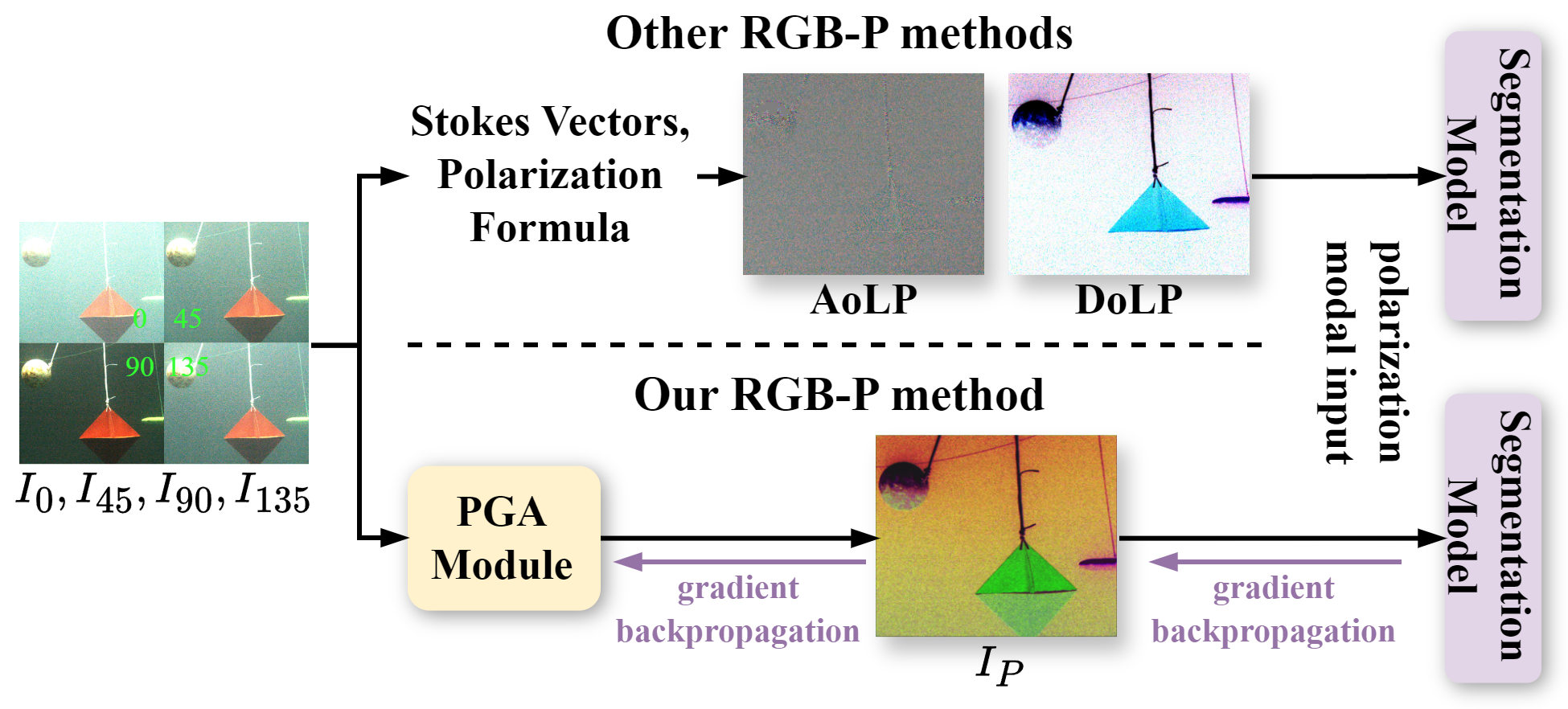}
\caption{Comparison between other RGB-P methods and our RGB-P method. Compared with other RGB-P methods that use fixed paradigms to generate polarization modal images (AoLP and DoLP), our method has higher adaptability. It dynamically generates polarization modal image inputs (\(I_{P}\)) by the PGA module and learns to make the generated polarization modal images more conducive to multimodal semantic segmentation models during training, thereby achieving full exploitation of polarization modal information.}
\label{fig:abstract}
\end{figure}

Existing approaches primarily utilize polarization information by calculating and representing the Degree of Linear Polarization (DoLP) and the Angle of Linear Polarization (AoLP) from polarized images as inputs for semantic segmentation~\cite{mcubes,2,zju,4,5}, as in Fig. \ref{fig:abstract}. While these approaches do not take into account the application of polarization information to underwater environments, and these representations capture essential polarization attributes, they reduce the rich polarization properties into fixed paradigms, potentially oversimplifying the potential of polarization modality. Thus, we argue that current methods underexplore the polarization properties, which may limit the effectiveness of polarization as a perceptual modality. This has been validated in our comparison experiment results (Tables \ref{tab:UPLight}, \ref{tab:ZJU}, and \ref{tab:MCubeS}).

Moreover, current multimodal semantic segmentation methods~\cite{28,29,31,32,66} designed for RGB-D and RGB-T, and mainstream methods~\cite{cmx,cmnext,32} (RGB-X) adopt a dual-branch architecture to extract features for RGB and X modalities separately and perform attention interaction and fusion for these two modalities. While these methods achieve strong performance on standard multimodal benchmarks, their dual-branch designs result in a large number of parameters, making them less suitable for resource-constrained (less running memory) AUV deployments. To overcome these challenges, we explore efficient architectures that maintain high performance while reducing parameters.

To delve into RGB-P multimodal semantic segmentation, we design a shared dual-branch encoder (ShareCMP Encoder) to reduce the number of parameters of the dual-branch model and reduce the use of running memory for model deployment. We investigate the performance of different polarization modalities in multimodal semantic segmentation and design a non-fixed paradigm Polarization Generate Attention (PGA) module using four-direction polarized images to generate the representation with richer polarization properties for the RGB-P multimodal semantic segmentation model. While based on the principle that the polarization angles of light reflected by different categories of materials are different, we propose a Class Polarization-Aware Loss (CPALoss) to assist the model in learning the polarization properties of different classes and improve the multimodal feature understanding and interaction capabilities of the RGB-P model. Meanwhile, we create an underwater RGB-P semantic segmentation benchmark that includes 0, 45, 90, and 135 degree polarized images with a total of 12 classes, which we call UPLight. Our method is called ShareCMP, inspired by CMX~\cite{cmx}. Unlike CMX, ShareCMP is a multimodal semantic segmentation model specifically designed for polarized images, with a ShareCMP Encoder, PGA, and CPALoss, which has better performance.

With extensive experiments on UPLight and two additional public RGB-P datasets, we obtain performance evaluations of the ShareCMP model. ShareCMP achieves top mIoU of 92.45{\small (+0.32)}\% on UPLight, 92.4{\small (+0.2)}\% (MiT-B2~\cite{26})/92.7{\small (+0.1)}\% (MiT-B4) on ZJU~\cite{zju}, and 50.34{\small (+1.92)}\% (RGB-A)/50.55{\small (+16.45)}\% (RGB-D)/50.99{\small (+1.51)}\% (RGB-A-D) on MCubeS~\cite{mcubes} datasets compared to the previous best methods. Our ShareCMP outperforms all previous RGB-P methods on these three datasets, achieving state-of-the-art performance with fewer parameters. And our ShareCMP (w/o PGA and CPALoss) achieves competitive or even higher performance on other RGB-X datasets compared to the corresponding state-of-the-art RGB-X methods.

In conclusion, we deliver the following contributions:
\begin{itemize}
    \item We consider the shortcomings of the current dual-branch multimodal model and present a ShareCMP Encoder to reduce the number of model parameters.
    \item We explore and compare different polarization modal representations and propose the PGA module to generate the representation with richer polarization properties.
    \item CPALoss is proposed, with Class Polarization-Aware Auxiliary Head (CPAAHead) to improve the learning and understanding of the encoder for polarization properties.
    \item We construct a new benchmark UPLight for underwater RGB-P semantic segmentation, which provides data support for AUVs to perform special perception tasks.
\end{itemize}

The rest of this paper is organized as follows: Section \ref{sec:related_work} reviews some related works. Section \ref{sec:method} describes the details of the proposed ShareCMP. Section \ref{sec:experiments} presents experimental results and analysis. Section \ref{sec:conclusion} presents conclusions and directions of future study.
\section{Related Work}
\label{sec:related_work}

\subsection{Unimodal Semantic Segmentation.}
Semantic segmentation can be seen as an extension of image classification from image level to pixel level. Since fully convolutional networks~\cite{6} introducing the end-to-end per-pixel classification paradigm, semantic segmentation has advanced significantly. The methods include capturing multi-scale features~\cite{7,8,9,63}, appending channel attention and self-attention blocks~\cite{11,12,13,60}, refining context priors~\cite{15,16,17,18,68}, and leveraging edge cues~\cite{19,20,61,22}. More recent methods prove the effectiveness of transformer-based architectures for semantic segmentation~\cite{23,62,25,26}. While these works achieve high performance, they still suffer under real-world conditions where RGB images do not provide sufficient textures such as low illumination. In underwater semantic segmentation, some methods improve semantic segmentation performance by using multi-scale features or attention~\cite{70,72,73}, or by incorporating denoising processes~\cite{71} into the model. However, these methods are still limited by the lack of sufficient textures in RGB images.

\subsection{Multimodal Semantic Segmentation.}
At present, multimodal semantic segmentation has been widely studied, such as RGB-Depth~\cite{27,28,29,64,dformer,geminifusion}, RGB-Thermal~\cite{mfnet,31,32,65,66,69,mdnet}, RGB-Event~\cite{33,eisnet}, RGB-LiDAR~\cite{34}, RGB-Polarization~\cite{zju,mcubes,2,4,5,40}, \textit{etc}. These methods adopt dual-branch or multi-branch architectures and use different attention modules between different branches for multi-scale cross-modal information interaction. Recently, unified multimodal segmentation CMX~\cite{cmx} and arbitrary-modal semantic segmentation CMNeXt~\cite{cmnext} have achieved better performance. Besides, a work~\cite{rgbpfusion} designs a fusion network with channel and spatial attention to construct underwater polarization fusion images to improve the performance of underwater vision tasks, such as semantic segmentation. It is not convenient for its fusion network and downstream task models to be separate. Currently, there is limited research on underwater RGB-P multimodal semantic segmentation, and there is also a lack of relevant underwater RGB-P datasets. However, the complex architecture of these methods, with their multiple branches and large number of parameters, presents significant challenges for efficient model deployment. While single-branch models based on data modal fusion~\cite{43} offer simpler solutions, their performance often falls short of that of dual-branch models. As such, there remains a pressing need to develop multimodal semantic segmentation methods that strike a balance between ease of use and high performance.

Among the diverse data modalities, the polarization modality presents unique challenges and opportunities. Unlike other modalities, the raw data for polarization consists of RGB images captured at different polarization angles, effectively forming a part of the RGB image. In this context, polarization can be considered a complementary layer of information embedded within the RGB data, which is fundamentally different from intensity-based modalities like depth or thermal. The traditional representations of polarization, such as the DoLP and AoLP~\cite{5,48}, are often calculated from the raw RGB-P data to extract meaningful features. However, we argue that these representations, while useful, fail to fully capture the underlying properties of the polarization data. This limitation arises from the oversimplification of the polarization process, which can only be partially represented by DoLP and AoLP. Polarized light carries richer information that can enhance the segmentation process, especially in underwater or low-light environments, where traditional methods often struggle.

The polarization modality involves not only the intensity but also the directional properties of the light, which provides additional layers of depth in understanding the scene. As a result, we believe that a more sophisticated treatment of polarization data, beyond DoLP and AoLP, is essential for improving the performance of RGB-P models. To address this, we propose ShareCMP, a shared dual-branch RGB-P multimodal semantic segmentation framework that aims to leverage the full potential of the polarization modality by exploiting more comprehensive polarization features, while also reducing the number of parameters in the network model.

\begin{figure*}[t]
\centering
\includegraphics[width=1.0\linewidth]{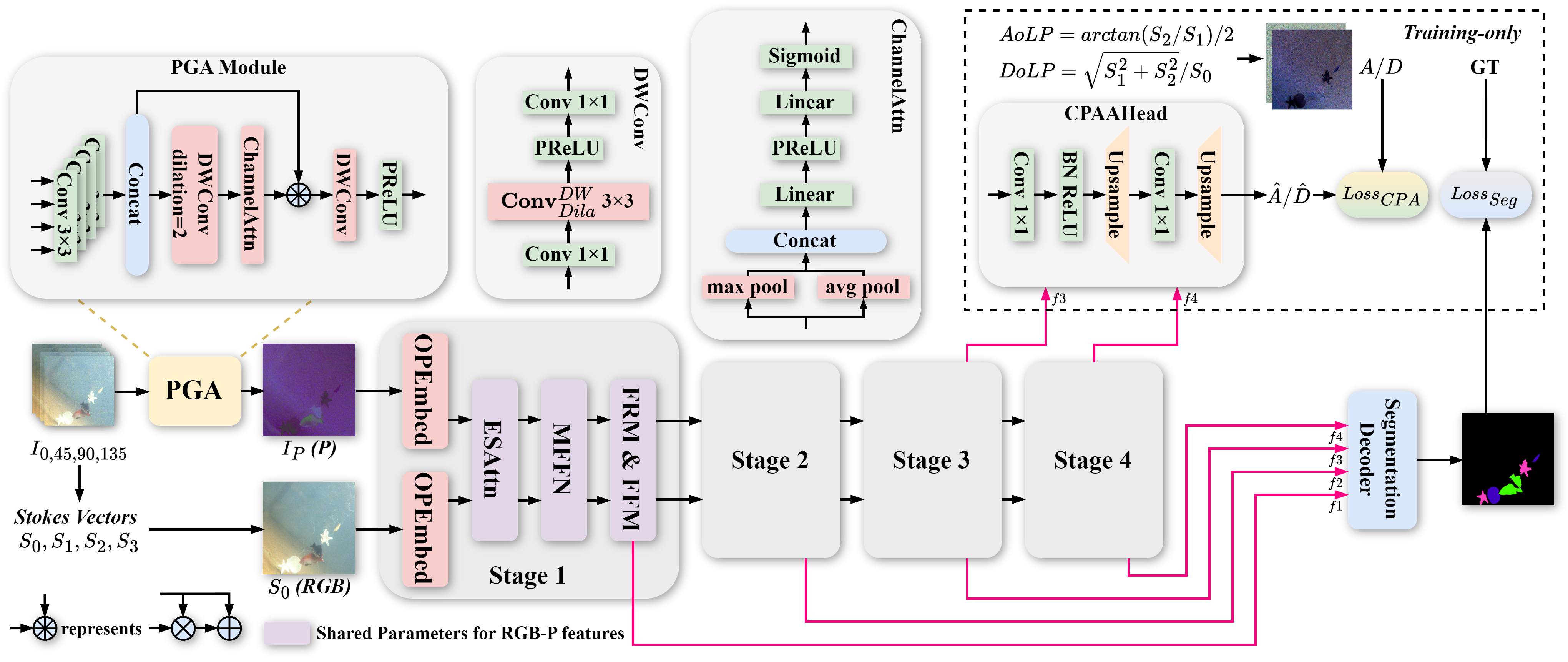}
\caption{ShareCMP framework. The ShareCMP Encoder consists of modules with modal exclusive Overlap Patch Embeddings (OPEmbed) and other modules that share parameters in each stage. The Polarization Generate Attention (PGA) module generates a polarization modal image input with rich polarization properties based on channel attention and large receptive fields. And the RGB input uses the \(S_{0}\) of Stokes vectors. The Class Polarization-Aware Auxiliary Head (CPAAHead) uses the \(\bm{f}_{3}\) and \(\bm{f}_{4}\) fused features in the ShareCMP encoder to construct Class Polarization-Aware Loss (CPALoss) \(Loss_{CPA}\) to improve the capability of the encoder to perceive optical polarization properties of different classes and optimize the backbone and PGA module. The architecture of the PGA module and CPAAHead is detailed above, with DWConv and ChannelAttn displayed on the right side of the PGA module (\(\mathrm{Conv}_{Dila}^{DW}\) represents depth-wise convolution with \textit{Dila} dilation). While the architecture of the OPEmbed, ESAttn, MFFN, FRM, and FFM are detailed in~\cite{26,cmx}.}
\label{fig:ShareCMP}
\end{figure*}
\section{Proposed Framework: ShareCMP}
\label{sec:method}

ShareCMP is designed based on CMX~\cite{cmx} and Segformer~\cite{26}. The Feature Rectification Module (FRM) and the Feature Fusion Module (FFM) from CMX, and the whole Segformer architecture are used in our ShareCMP. Besides, we design a shared dual-branch encoder, a Polarization Generate Attention Module (PGA) and a Class Polarization-Aware Loss (CPALoss). Four-direction polarized images (\(0^{\circ}\),\(45^{\circ}\),\(90^{\circ}\),\(135^{\circ}\)) are inputs for ShareCMP. And the output is the segmented image with object labels.

\subsection{ShareCMP Encoder}
\label{sec:encoder}

Both RGB and polarization modalities are encoded by a shared dual-branch encoder, and the four-stage architecture provides pyramidal fused features to the decoder. Meta-Transformer~\cite{metatransformer} shows that different modalities can be uniformly encoded using different data-to-sequence tokenization and a shared encoder. Inspired by this, we design a unified encoder with shared parameters for RGB and polarization modalities, as shown in Fig. \ref{fig:ShareCMP}. Each stage of the shared encoder consists of three modules: Overlap Patch Embeddings (OPEmbed), Efficient Self-Attention (ESAttn), and Mix-FFN (MFFN)~\cite{26}. OPEmbed generates feature patches with local continuous information. The feature patches are forwarded to the following ESAttn and MFFN modules. At each stage, the modal exclusive OPEmbed (ME OPEmbed) is constructed for the input \(\bm{x}_{RGB}\) and \(\bm{x}_{P}\), respectively, and using the shared-parameters ESAttn and MFFN introduces self-attention and leak location information~\cite{26} to obtain \(\bm{y}_{RGB}\) and \(\bm{y}_{P}\).
\begin{align}
    \label{eq:yrgbp}
    \bm{y}_{RGB}^{patch} & = \mathrm{OPEmbed}_{RGB}(\bm{x}_{RGB}) \\
    \label{eq:ypp}
    \bm{y}_{P}^{patch} & = \mathrm{OPEmbed}_{P}(\bm{x}_{P}) \\
    \label{eq:yrgb}
    \bm{y}_{RGB} & = \mathrm{MFFN}(\mathrm{ESAttn}(\bm{y}_{RGB}^{patch})) \\
    \label{eq:yp}
    \bm{y}_{P} & = \mathrm{MFFN}(\mathrm{ESAttn}(\bm{y}_{P}^{patch})) \\
    \bm{f} & = \mathrm{FFM}(\mathrm{FRM}(\bm{y}_{RGB},\bm{y}_{P}))
\end{align}
where \(\bm{y}_{X}^{patch}\) is feature patches of \(\bm{x}_{X}\). After that, the Feature Rectification Module (FRM) and the Feature Fusion Module (FFM)~\cite{cmx} introduce multimodal cross-attention for \(\bm{y}_{RGB}\) and \(\bm{y}_{P}\) and fuse their feature information. The FRM utilizes channel attention and spatial attention to fuse \(\bm{y}_{RGB}\) and \(\bm{y}_{P}\) and feeds the fused features of the two modalities into the next stage. The FFM utilizes multi-head attention to further fuse the two modal fusion features output by the FRM to obtain \(\bm{f}\), which is used for the final semantic segmentation. After the encoder, the four-stage fused features \(\bm{f}_{i} \in \{ \bm{f}_{1},\bm{f}_{2},\bm{f}_{3},\bm{f}_{4} \}\) are forwarded to the MLP decoder~\cite{26} for the segmentation prediction. We also refer to the shared dual-branch encoder (ShareCMP Encoder) as the pseudo-single-branch encoder. Although it still retains the dual-branch inference architecture, it reduces the number of multimodal encoder parameters by about 30\%.

\subsection{Polarization Generate Attention Module}
\label{sec:PGA}

To better represent the properties of polarization modal data, we develop the Polarization Generate Attention (PGA) module to generate polarization modal information with more polarization properties to better serve downstream tasks as semantic segmentation. As shown in Fig. \ref{fig:ShareCMP}, the PGA module uses Convolution (Conv) with \(3\times 3\) kernel size for four input images with 0, 45, 90, and 135 polarization angles \(\{ \bm{I}_{a} | a\in \{0,45,90,135\}, \bm{I}_{a} \in \mathbb{R}^{3 \times H \times W} \}\) to extract features (with \(d\) channels) and increases the receptive field of features without reducing the resolution of the image features. Finally, the concatenated feature \(\bm{f}_{P}\in \mathbb{R}^{4d \times H \times W}\) (\(4d\) is the number of channels of \(\bm{f}_{P}\), and the number of input and output channels for features in the subsequent network is \(4d\) in PGA) is obtained, as in Eq. \eqref{eq:fp}.
\begin{equation}
    \begin{split}
    \label{eq:fp}
    \bm{f}_{P} = \mathrm{Concat}( & \mathrm{Conv}_{3\times 3}(\bm{I}_{0}),
                                    \mathrm{Conv}_{3\times 3}(\bm{I}_{45}), \\
                                  & \mathrm{Conv}_{3\times 3}(\bm{I}_{90}),
                                    \mathrm{Conv}_{3\times 3}(\bm{I}_{135}) )
    \end{split}
\end{equation}
Previous multimodal fusion methods indicate channel information is crucial~\cite{46,47}. Inspired by this, after obtaining the concatenated feature \(\bm{f}_{P}\), using Depth-Wise Convolution (DWConv), which consists of convolution with \(1\times 1\) kernel size and with \(3\times 3\) kernel size composed of grouped convolution with 2 dilation~\cite{43}, provides the following channel attention module with advanced features with a larger receptive field to enhance the generated channel attention features \(\bm {Attn}_{P}\). ChannelAttn, which is optimized by the segmentation loss, weights the features of different angle-polarized images to achieve adaptive selection and enhancement of favorable polarization features. The processing can be formalized as:
\begin{equation}
    \label{eq:attnp}
    \bm{Attn}_{P} = \mathrm{ChannelAttn}(\mathrm{DWConv}_{dila=2}(\bm{f}_{P}))
\end{equation}
Finally, the concatenated feature \(\bm{f}_{P}\) is computed by attention, residual connection, feed-forward network composed of DWConv and PReLU~\cite{45} activation, then the three-channel polarization modal image \(\bm{I}_{P} \in \mathbb{R}^{3 \times H \times W}\) is generated. The DWConv is equivalent to a polarization modal feature decoder used to generate polarization modal images useful for segmentation tasks. It can be formalized as:
\begin{equation}
    \label{eq:P}
    \bm{I}_{P} = \mathrm{PReLU}(\mathrm{DWConv}(\bm{f}_{P} + \bm{Attn}_{P} * \bm{f}_{P}))
\end{equation}
The residual connection (shortcut) is to reduce the loss of raw polarization modal feature information caused by attention. The adaptive negative interval response of the activation function PReLU is more conducive to filtering and preserving polarization modal information~\cite{43}. The PGA module is a lightweight module with an architecture as in Fig. \ref{fig:ShareCMP}.

\subsection{Class Polarization-Aware Loss}
\label{sec:cpaloss}

To improve the learning and understanding of the encoder for polarization modal information, we design the Class Polarization-Aware Loss (CPALoss) and build the corresponding Class Polarization-Aware Auxiliary Head (CPAAHead) for CPALoss, as illustrated in Fig. \ref{fig:ShareCMP}. Based on the principle that the reflected light from different categories of materials produces different degrees of polarization and angles of polarization, CPALoss uses the estimates of AoLP and DoLP corresponding to different classes generated by CPAAHead and the ground truth AoLP and DoLP calculated from polarized images at four angles of 0, 45, 90, and 135 to calculate the loss \(Loss_{CPA}\). The loss is optimized to improve the capability of the encoder for perceiving the optical polarization properties of different classes while improving the quality of the polarization modal image generated by the PGA module. CPAAHead, which has a similar architecture to the MLP decoder~\cite{26}, uses the multi-scale multimodal fusion features \(\bm{f}_{i}\) generated in the encoder to estimate the AoLP and DoLP of different classes in the image, respectively. CPAAHead constructs two convolution layers with \(1\times 1\) kernel size for each scale feature. Upsampling operations are performed after the first convolution layer and the second convolution layer to unify the different scales of the features and to align the estimation and ground truth size of the AoLP or DoLP. The CPALoss can be written as:
\begin{align}
    \label{eq:AD}
    \hat{\bm{A}}_{i}/\hat{\bm{D}}_{i} & = \mathrm{Conv}^{up}_{1\times 1}(
                                          \mathrm{Conv}^{up}_{1\times 1}(\bm{f}_{i}) ) \\
    \label{eq:CPALoss}
    Loss_{CPA} & = \lambda\sum^{Sta}_{i}\sum^{Cls}_{c}(
    (\bm{A}^{c}_{i}-\hat{\bm{A}}^{c}_{i})^{2} + 
    (\bm{D}^{c}_{i}-\hat{\bm{D}}^{c}_{i})^{2} )
\end{align}
where \(i\in Sta \{ 1,2,3,4 \}\) represents the index of the \(i\) stage in the encoder, \(c\in Cls\) represents the class in the dataset. \(\mathrm{Conv}^{up}_{1\times 1}\) is a convolution with \(1\times 1\) followed by upsampling. \(\hat{\bm{A}}/\hat{\bm{D}}\), \(\bm{A}/\bm{D}\) are the estimation and ground truth of AoLP/DoLP, respectively. \(\lambda \) is the loss weight of CPALoss, which is set to 0.01 according to the ablation study in Section \ref{sec:ab_cpaloss}. We apply CPALoss at 3 and 4 encoder stages, which gives the best performance, a more detailed analysis is in Section \ref{sec:ablation_study}.
\section{Experiments}
\label{sec:experiments}

\begin{figure}[t]
\centering
\includegraphics[width=1.0\linewidth]{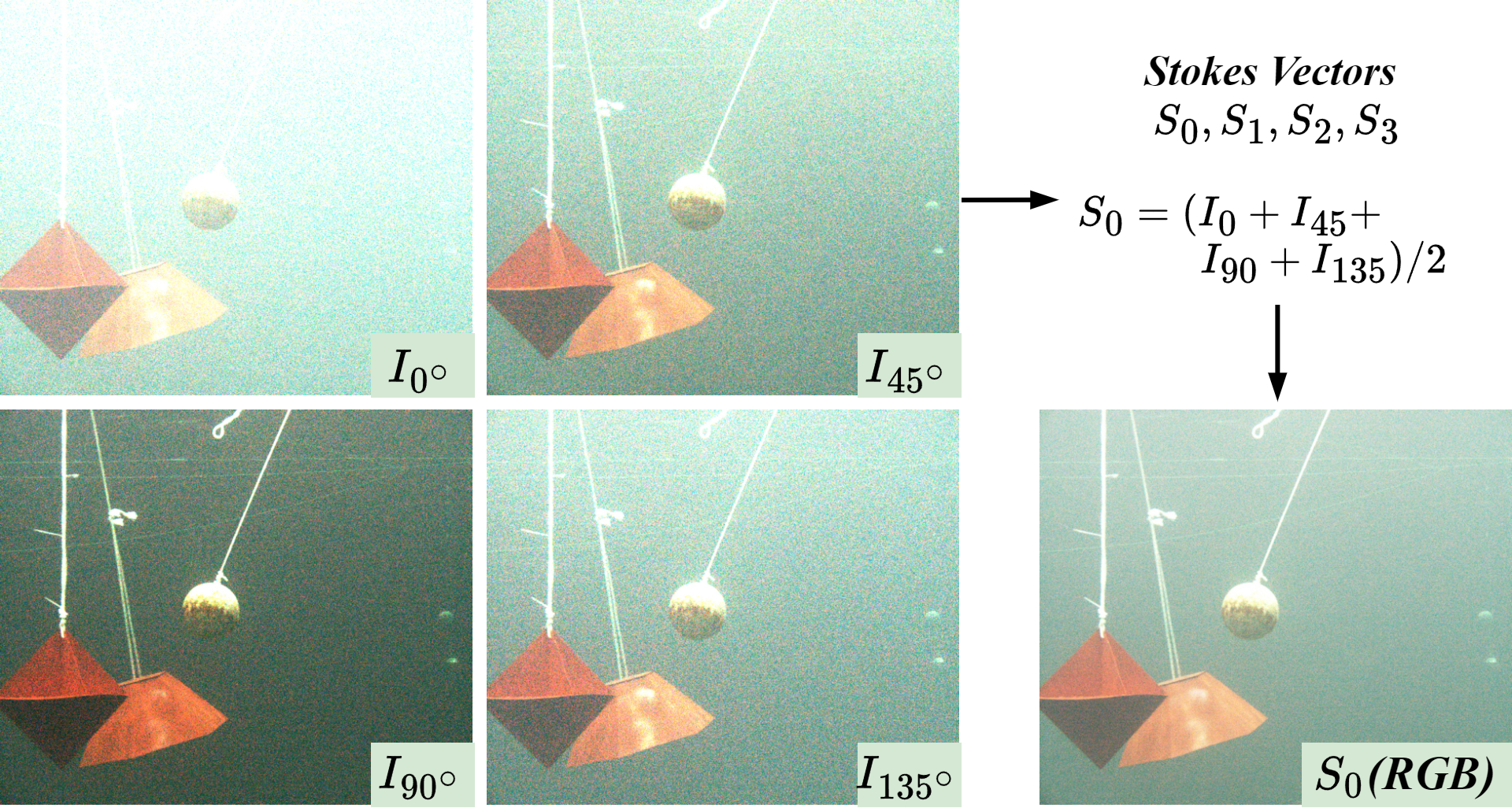}
\caption{An example image pair of our UPLight dataset including four-direction polarized images \(\bm{I}_{0^{\circ}},\bm{I}_{45^{\circ}},\bm{I}_{90^{\circ}},\bm{I}_{135^{\circ}}\) for the polarized modal image (P) and one total light intensity image \(S_{0}\) (RGB).}
\label{fig:UPLight_show}
\end{figure}

\subsection{Polarization Modal Representations}
\label{sec:pmodal_rep}

Polarization is a basic property of light, expressing the direction of light vibration. Polarized light can be categorized based on the nature of its vibration: linearly polarized light, circularly polarized light, and elliptically polarized light. Linearly polarized light, which is the focus of this study, exhibits vibration in a single, fixed direction or follows a regular change in direction. The polarization sensor is usually set four polarization plates with different angles (\(0^{\circ}\),\(45^{\circ}\),\(90^{\circ}\),\(135^{\circ}\)) on every four pixels, which collects linearly polarized images \{\(\bm{I}_{0}\),\(\bm {I}_{45}\),\(\bm{I}_{90}\),\(\bm{I}_{135}\)\} with pixels aligned at these four angles at the same time. We investigate four polarization modal representations~\cite{zju,5}, the Angle of Linear Polarization (AoLP), the Degree of Linear Polarization (DoLP), the Sin Angle of Linear Polarization (SAoLP), and the Cos Angle of Linear Polarization (CAoLP), which the SAoLP and CAoLP that we design are based on the principles of trigonometric functions. The SAoLP and CAoLP are equivalent to the decomposition of the AoLP. These polarization modal representations are derived from stokes vectors \(\bm{S} = \{ S_{0},S_{1},S_{2},S_{3} \}\) that describe the polarization state of the light. Precisely, \(S_{0}\) represents the total light intensity, \(S_{1}\) and \(S_{2}\) denote the ratio of \(0^{\circ}\) and \(45^{\circ}\) linear polarization over its perpendicular polarized portion, and \(S_{3}\) stands for the circular polarization power which is not involved in our work. The Stokes vectors \(\bm{S}\) can be calculated by the following Eq. \eqref{eq:S}, which provides a comprehensive description of the polarization properties of light based on the intensity and phase relationships of the light orthogonal components. This representation of polarization offers a richer understanding of the light field and enables a more detailed analysis of polarization in complex environments.
\begin{equation}
    \label{eq:S}
    \bm{S} = \left\{
    \begin{split}
    S_{0} & = \bm{I}_{0} + \bm{I}_{90}
            = \bm{I}_{45} + \bm{I}_{135} \\
    S_{1} & = \bm{I}_{0} - \bm{I}_{90} \\
    S_{2} & = \bm{I}_{45} - \bm{I}_{135}
    \end{split}
    \right.
\end{equation}

\begin{figure}[t]
\centering
\subfloat[]{
\includegraphics[width=0.30\linewidth]{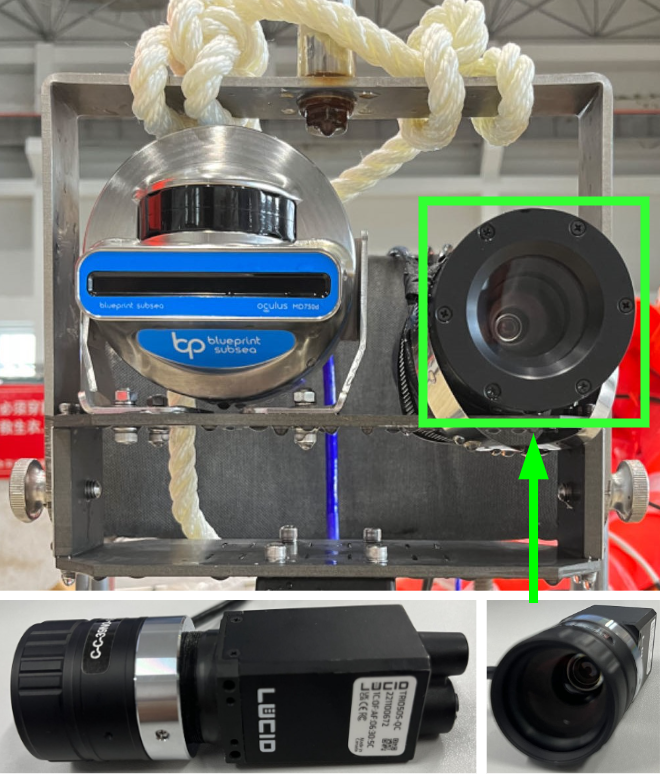}
\label{fig:UPLight_vis}
}
\hfill
\subfloat[]{
\includegraphics[width=0.62\linewidth]{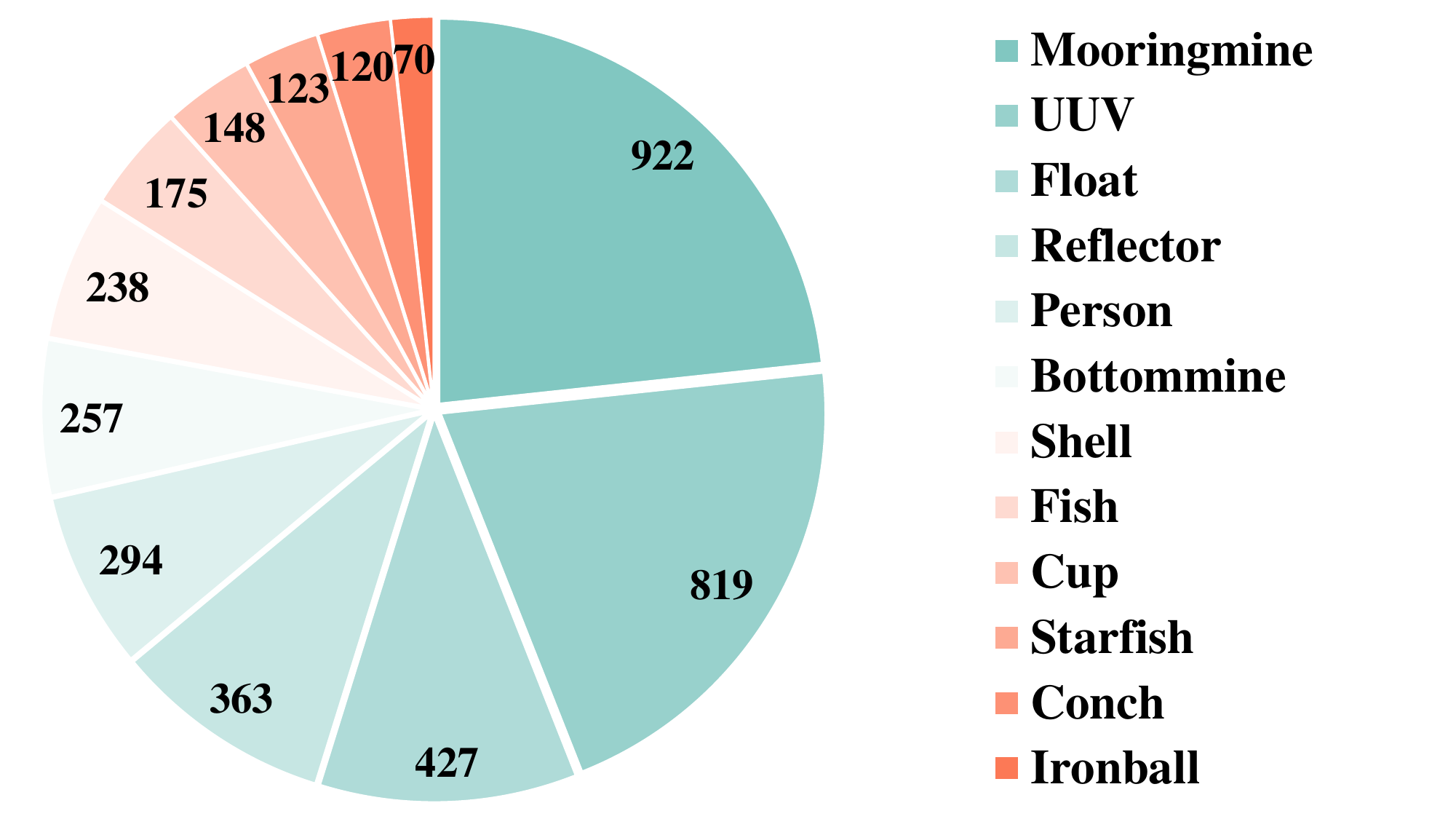}
\label{fig:UPLight_ana}
}
\caption{An introduction to our UPLight RGB-P multimodal semantic segmentation dataset. (a) Polarization camera for capturing the UPLight. The sonar on the left is used to assist us in locating the object. (b) Distribution of object quantity in 12 semantic classes.}
\label{fig:UPLight}
\end{figure}
\begin{figure}[t]
\centering
\includegraphics[width=1.0\linewidth]{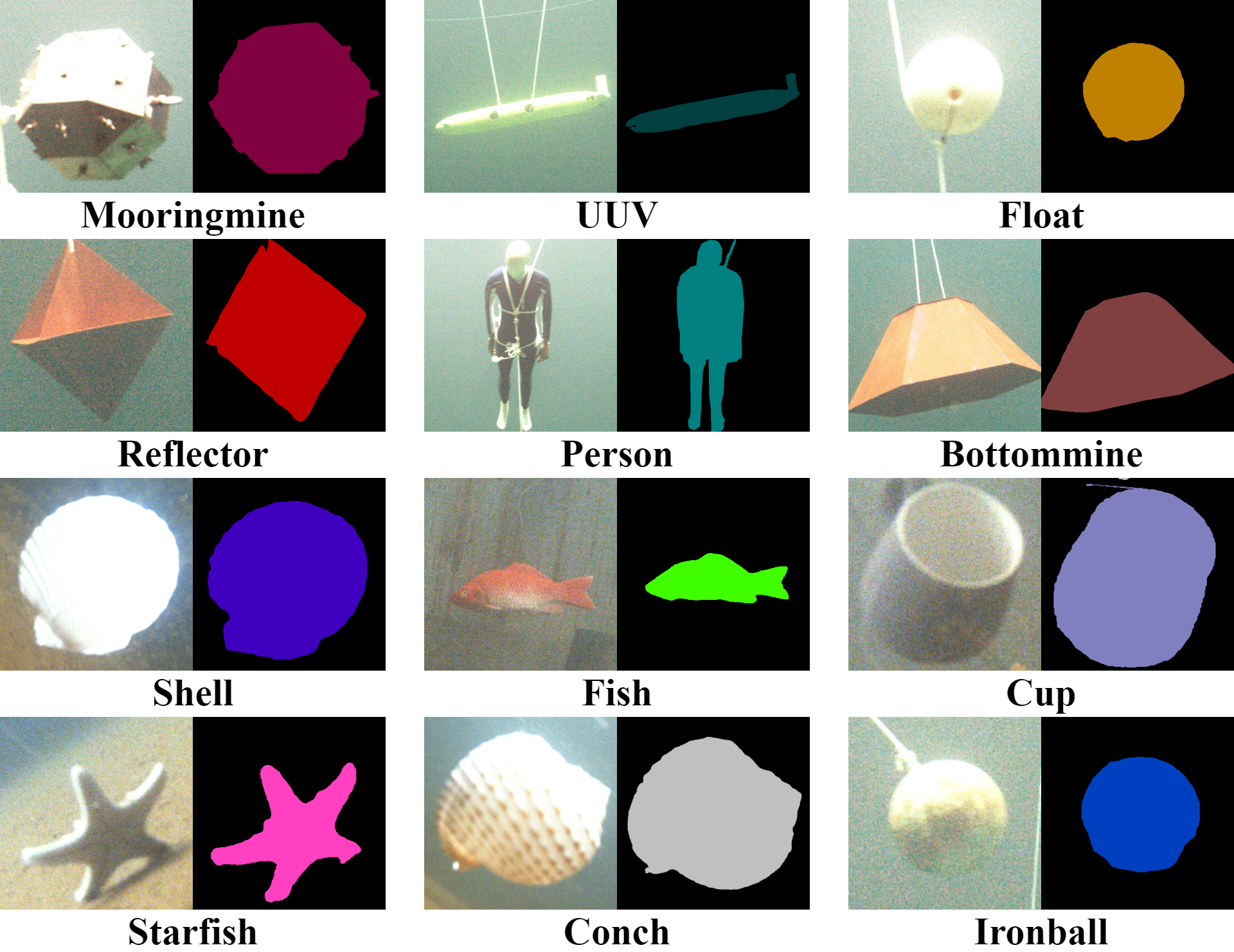}
\caption{12 semantic classes in our UPLight dataset. We annotate every pixel with 12 different classes and show example regions for each class, the annotated pixel contains different materials like metal, plastic, wood, calcium, and ceramic.}
\label{fig:UPLight_objs}
\end{figure}

In this work, to represent the total light intensity more accurately, we use Eq. \eqref{eq:S0} to calculate \(S_{0}\)~\cite{48,5}, and make \(S_{0}\) be the RGB input image of the RGB-P multimodal semantic segmentation model.
\begin{equation}
    \label{eq:S0}
    S_{0} = (\bm{I}_{0} + \bm{I}_{90} +
             \bm{I}_{45} + \bm{I}_{135}) / 2
\end{equation}
Then, AoLP, DoLP, SAoLP, and CAoLP are formally computed as follows:
\begin{align}
    \label{eq:AoLP}
    \mathrm{AoLP}  & = \frac{1}{2} \arctan(\frac{S_{2}}{S_{1}}) \\
    \label{eq:DoLP}
    \mathrm{DoLP}  & = \frac{ \sqrt{S^{2}_{1} + S^{2}_{2}} }{S_{0}} \\
    \label{eq:SAoLP}
    \mathrm{SAoLP} & = \frac{1}{2} \arcsin(\frac{S_{2}}{S_{0}}) \\
    \label{eq:CAoLP}
    \mathrm{CAoLP} & = \frac{1}{2} \arccos(\frac{S_{1}}{S_{0}})
\end{align}
Surfaces in the real-world made of different materials cause different subsurface compositions and surface structures at the mesoscopic scale. These differences give rise to different behaviors of incident light. Most notably, they alter the radiometric behavior of incident light in its reflection, refraction, and absorption. While it also conforms to the optical transmission model. These differences can often be observed in their polarization properties, and different degree of DoLP or AoLP. According to Eq. \eqref{eq:DoLP}, the DoLP ranges from 0 to 1. For partially polarized light, DoLP \(\in (0, 1)\), and for completely polarized light, DoLP \(=1\). The AoLP ranges from (\(0^{\circ}\) to \(180^{\circ}\)) and reflects silhouette information, as objects with similar materials often share similar AoLP. This makes AoLP a natural segmentation mask, grouping objects by material or category. Different from RGB-based sensors whose output can be influenced by various outdoor environments, polarization information from RGB-P sensor still remains stable according to Eq. \eqref{eq:S}, \eqref{eq:AoLP}, and \eqref{eq:DoLP}. Because the RGB images with four polarization directions from RGB-P sensor suffer similar degradation in the process of reaching the polarization filter in RGB sensor and the degradation will be cancelled out in the derivation of DoLP and AoLP. Therefore, the polarization stability against various environments is beneficial for scene perception.

\begin{table}[t]
\caption{Comparison of existing RGB-P semantic segmentation datasets.}
\label{tab:datasets}
\centering
\resizebox{\linewidth}{!}{

\begin{tabular}{@{}lccc@{}}
\toprule
Dataset & Modal & Scene & Total images \\ \midrule
ZJU~\cite{zju}          & RGB-P (\textit{FP})           & Street     & 394  \\
MCubeS~\cite{mcubes}    & RGB-P (\textit{A},\textit{D}) & Street     & 500  \\ \midrule
\textbf{UPLight} (Ours) & RGB-P (\textit{FP})           & Underwater & 1791 \\ \bottomrule
\end{tabular}

}
\end{table}

In our experiments, we further investigated the impact of AoLP, DoLP, SAoLP, and CAoLP on the performance of RGB-P multimodal semantic segmentation. Specifically, we used each of these polarization modal representations AoLP, DoLP, SAoLP, and CAoLP as individual inputs to the RGB-P model, allowing us to analyze and compare their effects on segmentation performance.

\subsection{Datasets and Implementation Details}
\label{sec:datasets_implementation}

\subsubsection{UPLight dataset} As in Table \ref{tab:datasets}, the current existing datasets are all for street scenes, and the dataset size is relatively small. The MCubeS dataset only contains AoLP and DoLP (\textit{A,D}) polarization modal images that are processed, instead ZJU dataset contains four raw polarized images (\textit{FP}) with \(0^{\circ}\),\(45^{\circ}\),\(90^{\circ}\),\(135^{\circ}\) polarization angles. We introduce the Underwater Polarization semantic segmentation dataset in the Light scene (UPLight). It is an underwater RGB-P semantic segmentation dataset that we create for AUV perception, which has 12 classes, Mooringmine, UUV, Float, Reflector, Person, Bottommine, Shell, Fish, Cup, Starfish, Conch, and Ironball. Fig. \ref{fig:UPLight_objs} shows examples of all semantic classes with their labels. Our image capture system consists of a RGB-P camera (LUCID\_TRI050S-QC, 2/3-in sensor, with a C-C-39N0-160-R12 liquid lens and customize a waterproof housing for underwater use, as the green box in Fig. \ref{fig:UPLight_vis}). The imaging system is fixed on the tank edge at a depth of 5 meters and collects data by moving floating objects in the tank. We mainly collect color-polarized images of typical underwater objects arranged in the tank to provide data support for AUV to perform perception tasks. The dataset is split into 1441/350 image pairs for training/validation at the size of \(1224\times 1024\) after data deduplication and filtering. The dataset is split according to the 8:2 ratio, where the number of semantic objects is evenly distributed in training and validation. And the distribution of 12 semantic classes in all images is shown in Fig. \ref{fig:UPLight_ana}. Each image pair includes one RGB image \(S_{0}\) and four polarized images with different polarization angles (\(0^{\circ}\),\(45^{\circ}\),\(90^{\circ}\),\(135^{\circ}\)). Fig. \ref{fig:UPLight_show} shows four pixel-aligned RGB images at four polarization directions and one total light intensity RGB image. The input image is resized to \(612\times 512\).

\subsubsection{ZJU~\cite{zju} dataset} It is an RGB-P multimodal dataset for automated driving on complex campus street scenes, which has 8 classes. It has 344/50 image pairs for training/validation at the size of \(1224\times 1024\). Each image pair includes four polarized images with different polarization angles (\(0^{\circ}\),\(45^{\circ}\),\(90^{\circ}\),\(135^{\circ}\)). The input image is resized to \(612\times 512\).

\subsubsection{MCubeS~\cite{mcubes} dataset} It is an RGB-P-NIR multimodal dataset for material segmentation on different street scenes, which has 20 classes. It has 302/96/102 image pairs for training/validation/test at the size of \(1224\times 1024\). Each image pair includes RGB, AoLP, DoLP, and Near-Infrared (NIR) images. The input image is resized to \(612\times 512\).

\subsubsection{NYU Depth V2~\cite{nyudepthv2} dataset} It is an RGB-D multimodal dataset for indoor understanding, which has 40 classes. It has 795/654 image pairs for training/test at the size of \(640\times 480\). The input image is resized to \(640\times 480\).

\subsubsection{SUN-RGBD~\cite{sunrgbd} dataset} It is an RGB-D multimodal dataset, which has 37 classes. It has 5285/5050 image pairs for training/test. The input image is resized to \(480\times 480\).

\subsubsection{ScanNetV2~\cite{scannetv2} dataset} It is an RGB-D multimodal dataset, which has 20 classes. It has 19466/5436/2135 image pairs for training/validation/test at the size of \(1296\times 968\). The input image is resized to \(640\times 480\).

\subsubsection{MFNet~\cite{mfnet} dataset} It is an RGB-T multimodal dataset, which has 8 classes. 820 images are captured during the day and the other 749 are at night. The training set has 50\% of the daytime and 50\% of the nighttime images, while the validation and test set respectively have 25\% of the daytime and 25\% of the nighttime images. The input image is resized to \(640\times 480\).

\subsubsection{EventScape~\cite{cmx} dataset} It is an RGB-E multimodal dataset, which has 12 classes. It has 4077/749 image pairs for training/test. The input image is resized to \(512\times 256\).

\subsubsection{KITTI-360~\cite{kitti360} dataset} It is an RGB-LiDAR multimodal dataset for suburban driving, which has 19 classes. It has 49004/12276 image pairs for training/validation. The input image is resized to \(1408\times 376\).

\subsubsection{Implementation details}
Our implementations are based on the MMSegmentation toolbox~\cite{mmseg}. Unless otherwise specified, we train our models on 2 A6000 GPUs with an initial learning rate (LR) of \(6e^{-5}\), which is scheduled by the poly strategy with power 1.0 over 200 epochs. The first 5 epochs are to warm up models with \(1e^{-6}\)\(\times\) the initial LR. The optimizer is AdamW~\cite{49} with a weight decay of 0.01, and the batch size is 4 per GPU. We use cross-entropy as the semantic segmentation loss function. The images are augmented by random resize with a ratio of 0.5-2.0, random horizontal flipping, random color jitter, and random cropping to \(612\times 512\) on RGB-P datasets. For other RGB-X datasets, the settings are as above. To conduct comparisons, the ImageNet-1K~\cite{50} pre-trained weight is used for ShareCMP on all datasets, but not for OPEmbed~\cite{26} modules in the polarization modal branch. We use mean Intersection over Union (mIoU) averaged across semantic classes as the primary evaluation metric to measure the segmentation performance.

\begin{table*}[t]
\caption{Results on UPLight.}
\label{tab:UPLight}
\centering
\resizebox{\linewidth}{!}{

\begin{tabular}{@{}lccccccccccccccc@{}}
\toprule
  Method &
  Modal &
  \rotatebox{90}{Unlabeled} &
  \rotatebox{90}{Mooringmine} &
  \rotatebox{90}{UUV} &
  \rotatebox{90}{Float} &
  \rotatebox{90}{Reflector} &
  \rotatebox{90}{Person} &
  \rotatebox{90}{Bottommine} &
  \rotatebox{90}{Cup} &
  \rotatebox{90}{Ironball} &
  \rotatebox{90}{Conch} &
  \rotatebox{90}{Fish} &
  \rotatebox{90}{Shell} &
  \rotatebox{90}{Starfish} &
  mIoU\(\uparrow\) \\ \midrule
  
SegFormer-B2~\cite{26}  & RGB      & 99.49 & 92.77 & 94.47 & 89.12 & 97.96 & 95.00 & 97.44 & 89.66 & 82.92 & 84.12 & 89.31 & 83.25 & 69.25 & 89.60 \\ \midrule
MCubeSNet~\cite{mcubes} & RGB-AoLP & 89.72 & 83.32 & 84.78 & 85.20 & 88.79 & 85.63 & 88.94 & 81.76 & 75.47 & 83.37 & 80.77 & 81.40 & 65.18 & 82.64 \\
EAFNet~\cite{zju}       & RGB-AoLP & 94.61 & 88.21 & 89.67 & 90.09 & 93.68 & 90.52 & 93.83 & 86.65 & 80.36 & 88.26 & 85.66 & 86.29 & 70.07 & 87.53 \\
CMX (MiT-B2)~\cite{cmx} & RGB-AoLP & 99.53 & 93.08 & \textbf{94.90} & 92.27 & 98.07 & 95.27 & 97.88 & 88.67 & 84.83 & \textbf{93.60} & 90.40 & \textbf{92.62} & \textbf{76.53} & 92.13 \\ \midrule
MCubeSNet               & RGB-DoLP & 87.88 & 81.48 & 82.94 & 83.36 & 86.95 & 83.79 & 87.10 & 79.92 & 73.63 & 81.53 & 78.93 & 79.56 & 63.34 & 80.80 \\
EAFNet                  & RGB-DoLP & 94.42 & 88.02 & 89.48 & 89.90 & 93.49 & 90.33 & 93.64 & 86.46 & 80.17 & 88.07 & 85.47 & 86.10 & 69.88 & 87.34 \\
CMX (MiT-B2)            & RGB-DoLP & 99.53 & 92.88 & 94.79 & 90.48 & 98.19 & 95.22 & 97.51 & \textbf{92.00} & \textbf{85.48} & 91.26 & \textbf{90.95} & 92.51 & 76.17 & 92.07 \\ \midrule
ShareCMP (MiT-B2)       & RGB-\{\(\bm{I}_{0}\),\(\bm{I}_{45}\),\(\bm{I}_{90}\),\(\bm{I}_{135}\)\} & \textbf{99.53} & \textbf{93.13} & 94.59 & \textbf{95.01} & \textbf{98.60} & \textbf{95.44} & \textbf{98.75} & 91.57 & 85.28 & 93.18 & 90.58 & 91.21 & 74.99 & \textbf{92.45} \\ \bottomrule
\end{tabular}

}
\end{table*}
\begin{table*}[t]
\caption{Results on ZJU~\cite{zju}. ShareCMP is trained for 500 epochs.}
\label{tab:ZJU}
\centering
\resizebox{\linewidth}{!}{

\begin{tabular}{@{}lcccccccccc@{}}
\toprule
Method                  & Modal    & Building & Glass & Car   & Road  & Tree  & Sky   & Pedestrian & Bicycle & mIoU\(\uparrow\) \\ \midrule
SwiftNet~\cite{54}      & RGB      & 83.0     & 73.4  & 91.6  & 96.7  & 94.5  & 84.7  & 36.1       & 82.5    & 80.3 \\
SegFormer-B2~\cite{26}  & RGB      & 90.6     & 79.0  & 92.8  & 96.6  & 96.2  & 89.6  & 82.9       & 89.3    & 89.6 \\ \midrule
NLFNet~\cite{4}         & RGB-AoLP & 85.4     & 77.1  & 93.5  & 97.7  & 93.2  & 85.9  & 56.9       & 85.5    & 84.4 \\
EAFNet~\cite{zju}       & RGB-AoLP & 87.0     & 79.3  & 93.6  & 97.4  & 95.3  & 87.1  & 60.4       & 85.6    & 85.7 \\
EAFNet                  & RGB-DoLP & 86.4     & 76.9  & 93.0  & 97.1  & 95.5  & 86.1  & 62.1       & 86.0    & 85.4 \\
CMX (MiT-B2)~\cite{cmx} & RGB-AoLP & 91.5     & 87.3  & 95.8  & 98.2  & 96.6  & 89.3  & 85.6       & 91.9    & 92.0 \\
CMX (MiT-B2)            & RGB-DoLP & 91.8     & 87.8  & 96.1  & 98.2  & 96.7  & 89.4  & 86.1       & 91.8    & 92.2 \\
ShareCMP (MiT-B2)       & RGB-\{\(\bm{I}_{0}\),\(\bm{I}_{45}\),\(\bm{I}_{90}\),\(\bm{I}_{135}\)\}    & 91.4     & 88.4  & 96.2  & 98.2  & 96.7  & \textbf{90.2}  & 85.7       & 92.2    & 92.4 \\ \midrule
CMX (MiT-B4)            & RGB-AoLP & 91.6     & \textbf{88.8}  & 96.3  & 98.3  & 96.8  & 89.7  & 86.2       & \textbf{92.8}    & 92.6 \\
CMX (MiT-B4)            & RGB-DoLP & 91.6     & 88.6  & 96.3  & 98.3  & 96.7  & 89.5  & 86.4       & 92.2    & 92.5 \\
ShareCMP (MiT-B4)       & RGB-\{\(\bm{I}_{0}\),\(\bm{I}_{45}\),\(\bm{I}_{90}\),\(\bm{I}_{135}\)\}    & \textbf{92.0}     & 88.6  & \textbf{96.3}  & \textbf{98.3}  & \textbf{96.8}  & 90.0  & \textbf{87.2}       & 92.5    & \textbf{92.7} \\ \bottomrule
\end{tabular}

}
\end{table*}
\begin{table*}[t]
\caption{Results on MCubeS~\cite{mcubes}. The human body class is omitted as its result is 0\%.}
\label{tab:MCubeS}
\centering
\resizebox{\linewidth}{!}{

\begin{tabular}{@{}lccccccccccccccccccccc@{}}
\toprule
  Method &
  Modal &
  \rotatebox{90}{Asphalt} &
  \rotatebox{90}{Concrete} &
  \rotatebox{90}{Metal} &
  \rotatebox{90}{Road Ma} &
  \rotatebox{90}{Fabric} &
  \rotatebox{90}{Glass} &
  \rotatebox{90}{Plaster} &
  \rotatebox{90}{Plastic} &
  \rotatebox{90}{Rubber} &
  \rotatebox{90}{Sand} &
  \rotatebox{90}{Gravel} &
  \rotatebox{90}{Ceramic} &
  \rotatebox{90}{Cobbles} &
  \rotatebox{90}{Brick} &
  \rotatebox{90}{Grass} &
  \rotatebox{90}{Wood} &
  \rotatebox{90}{Leaf} &
  \rotatebox{90}{Water} &
  \rotatebox{90}{Sky} &
  mIoU\(\uparrow\) \\ \midrule

DRConv~\cite{55}              & RGB-A-D-N & -    & -    & -    & -    & -    & -    & -   & -    & -    & -    & -    & -    & -    & -    & -    & -    & -    & -    & -    & 34.63 \\
DDF~\cite{56}                 & RGB-A-D-N & -    & -    & -    & -    & -    & -    & -   & -    & -    & -    & -    & -    & -    & -    & -    & -    & -    & -    & -    & 36.16 \\
TransFuser~\cite{57}          & RGB-A-D-N & -    & -    & -    & -    & -    & -    & -   & -    & -    & -    & -    & -    & -    & -    & -    & -    & -    & -    & -    & 37.66 \\
MMTM~\cite{58}                & RGB-A-D-N & -    & -    & -    & -    & -    & -    & -   & -    & -    & -    & -    & -    & -    & -    & -    & -    & -    & -    & -    & 39.71 \\
FuseNet~\cite{59}             & RGB-A-D-N & -    & -    & -    & -    & -    & -    & -   & -    & -    & -    & -    & -    & -    & -    & -    & -    & -    & -    & -    & 40.58 \\
MCubeSNet~\cite{mcubes}       & RGB-A-D-N & 85.7 & 42.6 & 47.0 & 59.2 & 12.5 & 44.3 & 3.0 & 10.6 & 12.7 & 66.8 & 67.1 & 27.8 & 65.8 & 36.8 & 54.8 & 39.4 & 73.0 & 13.3 & 94.8 & \textbf{42.90} \\ \midrule
MCubeSNet                     & RGB-A     & 83.3 & 42.3 & 43.0 & 58.4 & 8.8  & 27.3 & 0.6 & 9.8  & 12.0 & 55.5 & 57.7 & 18.1 & 64.6 & 36.6 & 56.5 & 34.8 & 71.8 & 6.8  & 95.0 & 39.10 \\
CMNeXt (MiT-B2)~\cite{cmnext} & RGB-A     & -    & -    & -    & -    & -    & -    & -   & -    & -    & -    & -    & -    & -    & -    & -    & -    & -    & -    & -    & 48.42 \\
ShareCMP (MiT-B2)             & RGB-A     & \textbf{88.2} & \textbf{49.3} & \textbf{51.0} & \textbf{66.1} & \textbf{20.7} & \textbf{50.1} & \textbf{1.2} & \textbf{30.9} & \textbf{18.2} & \textbf{64.2} & \textbf{75.2} & \textbf{28.5} & \textbf{74.9} & \textbf{45.7} & \textbf{59.4} & \textbf{43.9} & \textbf{74.1} & \textbf{69.9} & \textbf{95.6} & \textbf{50.34} \\ \midrule
MCubeSNet                     & RGB-D     & 75.2 & 40.2 & 37.8 & 53.9 & 4.2  & 32.3 & \textbf{1.9} & 14.3 & 11.3 & 59.7 & 21.8 & 11.6 & 28.9 & 29.1 & 54.6 & 29.4 & 71.4 & 9.6  & 94.3 & 34.10 \\
ShareCMP (MiT-B2)             & RGB-D     & \textbf{88.5} & \textbf{50.4} & \textbf{52.6} & \textbf{65.9} & \textbf{23.2} & \textbf{49.8} & 0.0 & \textbf{36.7} & \textbf{18.8} & \textbf{64.6} & \textbf{75.4} & \textbf{30.4} & \textbf{72.2} & \textbf{45.3} & \textbf{57.3} & \textbf{43.9} & \textbf{74.5} & \textbf{66.1} & \textbf{95.7} & \textbf{50.55} \\ \midrule
MCubeSNet                     & RGB-A-D   & 83.0 & 42.6 & 45.5 & 59.8 & 17.0 & 44.2 & \textbf{1.2} & 18.6 & 4.8  & 54.8 & 51.5 & 26.4 & 67.6 & 41.9 & 57.0 & 39.4 & 74.0 & 15.5 & 95.3 & 42.00 \\
CMNeXt (MiT-B2)               & RGB-A-D   & -    & -    & -    & -    & -    & -    & -   & -    & -    & -    & -    & -    & -    & -    & -    & -    & -    & -    & -    & 49.48 \\
ShareCMP (MiT-B2)             & RGB-A-D   & \textbf{88.8} & \textbf{49.7} & \textbf{52.5} & \textbf{66.4} & \textbf{27.6} & \textbf{51.0} & 0.2 & \textbf{31.5} & \textbf{18.0} & \textbf{69.9} & \textbf{79.0} & \textbf{30.9} & \textbf{71.4} & \textbf{42.8} & \textbf{58.3} & \textbf{44.1} & \textbf{74.5} & \textbf{67.6} & \textbf{95.6} & \textbf{50.99} \\ \bottomrule
\end{tabular}

}
\end{table*}
\begin{table*}[t]
\centering
\caption{The results on other RGB-X datasets. ShareCMP* represents ShareCMP w/o PGA and CPALoss for non-RGB-P datasets.
(a) Results on NYU Depth V2~\cite{nyudepthv2}.
(b) Results on SUN-RGBD~\cite{sunrgbd}.
(c) Results on MFNet~\cite{mfnet}.
(d) Results on ScanNetV2~\cite{scannetv2}.
(e) Results on EventScape~\cite{cmx}.
(f) Results on KITTI-360~\cite{kitti360}.
}
\label{tab:rgbx}

\subfloat[\label{tab:nyudepthv2}]{
\resizebox{0.3\linewidth}{!}{

\begin{tabular}{@{}lcc@{}}
\toprule
Method    & Modal & mIoU\(\uparrow\) \\ \midrule
RDF-101~\cite{rdf}   & RGB-D & 49.1 \\
SGNet~\cite{sgnet}     & RGB-D & 51.1 \\
ShapeConv~\cite{shapeconv} & RGB-D & 51.3 \\
NANet~\cite{nanet}     & RGB-D & 52.3 \\
SA-Gate~\cite{46}   & RGB-D & 52.4 \\
CMX~\cite{cmx}       & RGB-D & 54.1 \\ \midrule
ShareCMP* & RGB-D & \textbf{54.1} \\ \bottomrule
\end{tabular}

}
}
\hfill
\subfloat[\label{tab:sunrgbd}]{
\resizebox{0.3\linewidth}{!}{

\begin{tabular}{@{}lcc@{}}
\toprule
Method    & Modal & mIoU\(\uparrow\) \\ \midrule
RDF-152~\cite{rdf}   & RGB-D & 47.7 \\
TCD~\cite{tcd}       & RGB-D & 49.5 \\
SGNet~\cite{sgnet}     & RGB-D & 48.6 \\
SA-Gate~\cite{46}   & RGB-D & 49.4 \\
NANet~\cite{nanet}     & RGB-D & 48.8 \\
ShapeConv~\cite{shapeconv} & RGB-D & 48.6 \\
CMX~\cite{cmx}       & RGB-D & 49.7 \\ \midrule
ShareCMP* & RGB-D & \textbf{49.7} \\ \bottomrule
\end{tabular}

}
}
\hfill
\subfloat[\label{tab:mfnet}]{
\resizebox{0.3\linewidth}{!}{

\begin{tabular}{@{}lcc@{}}
\toprule
Method    & Modal & mIoU\(\uparrow\) \\ \midrule
HRNet~\cite{hrnet}     & RGB   & 51.7 \\
SegFormer~\cite{26} & RGB   & 53.2 \\ \midrule
ACNet~\cite{47}     & RGB-T & 46.3 \\
FuseSeg~\cite{fuseseg}   & RGB-T & 54.5 \\
ABMDRNet~\cite{31}  & RGB-T & 54.8 \\
LASNet~\cite{69}    & RGB-T & 54.9 \\
FEANet~\cite{feanet}    & RGB-T & 55.3 \\
MFTNet~\cite{mftnet}    & RGB-T & 57.3 \\
GMNet~\cite{65}     & RGB-T & 57.3 \\
CMX~\cite{cmx}      & RGB-T & 58.2 \\ \midrule
ShareCMP* & RGB-T & \textbf{58.3} \\ \bottomrule
\end{tabular}

}
}
\vspace{-12mm}
\subfloat[\label{tab:scannetv2}]{
\resizebox{0.3\linewidth}{!}{

\begin{tabular}{@{}lcc@{}}
\toprule
Method    & Modal & mIoU\(\uparrow\) \\ \midrule
PSPNet~\cite{10}    & RGB   & 47.5 \\
AdapNet++~\cite{ssma} & RGB   & 50.3 \\ \midrule
FuseNet~\cite{59}   & RGB-D & 53.5 \\
SSMA~\cite{ssma}      & RGB-D & 57.7 \\
GRBNet~\cite{28}    & RGB-D & 59.2 \\
MCA-Net~\cite{mcanet}   & RGB-D & 59.5 \\
DMMF~\cite{dmmf}      & RGB-D & 59.7 \\
CMX~\cite{cmx}       & RGB-D & 61.3 \\ \midrule
ShareCMP* & RGB-D & \textbf{61.4} \\ \bottomrule
\end{tabular}

}
}
\hfill
\raisebox{-3mm}{
\subfloat[\label{tab:eventscape}]{
\resizebox{0.3\linewidth}{!}{

\begin{tabular}{@{}lcc@{}}
\toprule
Method    & Modal & mIoU\(\uparrow\) \\ \midrule
SwiftNet~\cite{54}  & RGB   & 36.7 \\
CGNet~\cite{cgnet}     & RGB   & 44.8 \\
SegFormer~\cite{26} & RGB   & 58.7 \\ \midrule
RFNet~\cite{rfnet}     & RGB-E & 41.3 \\
ISSAFE~\cite{33}    & RGB-E & 43.6 \\
SA-Gate~\cite{46}   & RGB-E & 53.9 \\
CMX~\cite{cmx}       & RGB-E & 61.9 \\ \midrule
ShareCMP* & RGB-E & \textbf{62.0} \\ \bottomrule
\end{tabular}

}
}
}
\hfill
\raisebox{-19mm}{
\subfloat[\label{tab:kitti360}]{
\resizebox{0.3\linewidth}{!}{

\begin{tabular}{@{}lcc@{}}
\toprule
Method      & Modal     & mIoU\(\uparrow\) \\ \midrule
HRFuser~\cite{hrfuser}     & RGB-LiDAR & 48.7 \\
PMF~\cite{34}         & RGB-LiDAR & 54.5 \\
TokenFusion~\cite{tokenfusion} & RGB-LiDAR & 54.6 \\
TransFuser~\cite{57}  & RGB-LiDAR & 56.6 \\
CMX~\cite{cmx}         & RGB-LiDAR & 64.3 \\ \midrule
ShareCMP*   & RGB-LiDAR & \textbf{64.5} \\ \bottomrule
\end{tabular}

}
}
}

\end{table*}

\subsection{Comparison against State-of-the-arts}
\label{sec:comparison_sota}

To verify the efficacy of our proposed ShareCMP framework, we perform extensive experiments on three RGB-P segmentation datasets and other RGB-X segmentation datasets. The results and comparisons against the state-of-the-art are shown in Tables \ref{tab:UPLight}, \ref{tab:ZJU}, \ref{tab:MCubeS}, and \ref{tab:rgbx}.

\subsubsection{Results on UPLight}
As shown in Table \ref{tab:UPLight}, we train models~\cite{zju,mcubes,cmx} that can be used for RGB-P semantic segmentation and compare them with ShareCMP on the UPLight dataset we built. Our ShareCMP achieves the best performance of 92.45\%. Compared to the RGB-only baseline with SegFormer-B2~\cite{26}, the IoU improvements on classes with polarization properties are clear, such as Float (+5.9\%), Bottom mine (+1.3\%), Ironball (+2.4\%), Conch (+9.1\%), Shell (+8.0\%), and Starfish (+5.7\%). This also shows that our RGB-P multimodal semantic segmentation model ShareCMP has important application significance in underwater aquaculture, fishing, ocean exploration, \textit{etc}. Among them, Ironball and Starfish have an IoU below 90\% compared to other categories. We analyze it mainly because the starfish we used are dried starfish specimens, and there is a lot of rust attached to the surface of the iron ball. The surface roughness of both is very high, causing diffuse reflection of light and reducing the optical information collected by the polarization camera, resulting in their IoU being lower than 90\%. Due to the more complex surface structure of Starfish, its IoU is lower than that of Ironball. ShareCMP has only about 66.2\% of the number of parameters of CMX~\cite{cmx} (detailed parameter number analysis in Section \ref{sec:qv_analysis}), and has a 0.32\% performance improvement compared to CMX, which provides more advantages in the deployment of ShareCMP and its performance improvement with fewer parameters on AUVs.

\subsubsection{Results on ZJU}
In Table \ref{tab:ZJU}, we perform experiments to compare the ShareCMP method with the previous state-of-the-art methods~\cite{54,4,zju,cmx} on the ZJU~\cite{zju} dataset. Our ShareCMP outperforms the previous best RGB-P method by 0.2\% and 0.1\% based on MiT-B2~\cite{26} and MiT-B4, respectively. Compared to dual-branch multimodal semantic segmentation models~\cite{4,zju,cmx}, our ShareCMP beats them in segmentation performance with fewer parameters.

\subsubsection{Results on MCubeS}
In Table \ref{tab:MCubeS}, we benchmark 7 semantic segmentation methods on the MCubeS~\cite{mcubes} dataset, which contains AoLP, DoLP, and NIR modalities. Since the MCubeS dataset does not provide corresponding four-direction polarized images, our ShareCMP uses AoLP and DoLP as polarization modal inputs. On RGB-A and RGB-D modal inputs, our ShareCMP outperforms the previous best method by 1.92\% on CMNeXt~\cite{cmnext} and 16.45\% on MCubeSNet~\cite{mcubes}. ShareCMP achieves a state-of-the-art performance of 50.99\% when concatenating AoLP and DoLP modalities. In addition, our ShareCMP (RGB-A-D) even outperforms previous methods~\cite{55,56,57,58,59,mcubes} using RGB-A-D-N modalities, by 8.09\% compared to the best. The results indicate that our ShareCMP has great potential in RGB-P multimodal semantic segmentation.

\begin{table}[t]
\caption{Ablation study of modal exclusive OPEmbed (ME OPEmbed) on different encoder stages.}
\label{tab:OPEmbed}
\centering

\begin{tabular}{@{}cccccc@{}}
\toprule
\multirow{2}{*}{ShareCMP} & \multicolumn{4}{c}{Stage} & \multirow{2}{*}{mIoU\(\uparrow\)} \\ \cmidrule(lr){2-5}
              & 1          & 2          & 3          & 4          &                \\ \midrule
w/ ME OPEmbed & \checkmark &            &            &            & 91.91          \\
w/ ME OPEmbed & \checkmark & \checkmark &            &            & 92.36          \\
w/ ME OPEmbed & \checkmark & \checkmark & \checkmark &            & 92.01          \\
w/ ME OPEmbed & \checkmark & \checkmark & \checkmark & \checkmark & \textbf{92.45} \\ \bottomrule
\end{tabular}

\end{table}
\begin{table}[t]
\caption{Ablation study of PGA architecture settings.}
\label{tab:PGA_arch}
\centering
\resizebox{\linewidth}{!}{

\begin{tabular}{@{}lccc@{}}
\toprule
Method & \#Params (M)\(\downarrow\) & FLOPs (G)\(\downarrow\) & mIoU\(\uparrow\) \\ \midrule
PGA                      & 0.16 & 17.37 & 92.45 \\ \midrule
PGA w/o ChannelAttn      & 0.06 & 17.33 & 90.93 \\
PGA w/o DWConv (w/ Conv) & 0.40 & 93.20 & \textbf{92.46} \\
PGA w/o PReLU (w/ ReLU)  & 0.16 & 17.37 & 91.85 \\
PGA w/o shortcut         & 0.16 & 17.37 & 92.03 \\ \midrule
PGA w/ dilation=1        & 0.16 & 17.37 & 92.21 \\
PGA w/ dilation=3        & 0.16 & 17.37 & 92.01 \\ \bottomrule
\end{tabular}

}
\end{table}

\subsubsection{Results on Other RGB-X Datasets}
As shown in Table \ref{tab:rgbx}, we compare our ShareCMP without PGA module and CPALoss (ShareCMP*) on RGB-D, RGB-T, RGB-E, and RGB-LiDAR datasets. ShareCMP* achieves competitive or even higher performance on these datasets compared to the corresponding state-of-the-art RGB-X methods, it indicates that the effectiveness of our ShareCMP framework with the shared encoder in the RGB-X semantic segmentation task.

\subsection{Ablation Study}
\label{sec:ablation_study}

\subsubsection{Ablation of ShareCMP Encoder}
The ablation of setting non-shared parameters modal exclusive OPEmbed~\cite{26} on different stages of ShareCMP encoder is conducted, as shown in Table \ref{tab:OPEmbed}. We added ME OPEmbed from stage 1 to 4 because the RGB-P modalities have significant differences in the low-level stage, making it difficult to extract their features using a fully shared encoder (similar to multitasking problems, here fully shared network modules will cause the model to converge to a better local optimal solution for a certain modality). Therefore, ME OPEmbed must be present in the low-level stage in the ablation experiment. When the ShareCMP encoder sets modal exclusive OPEmbed in stage 1, its encoder architecture is similar to that of the Meta-Transformer unified encoder~\cite{metatransformer}, but it has poor performance. We believe that ShareCMP has multimodal attention interaction operations (by FRM and FFM~\cite{cmx}) after each stage. Simply setting a modal exclusive OPEmbed in stage 1 cannot meet the generalization performance requirements of the ShareCMP encoder. The results show that ShareCMP achieves the best performance when modal exclusive OPEmbed is set in all encoder stages.

\begin{table}[t]
\caption{Ablation study of the channels per polarized image feature in the PGA.}
\label{tab:PGA_channels}
\centering

\begin{tabular}{@{}cccc@{}}
\toprule
Channels & \#Params (M)\(\downarrow\) & FLOPs (G)\(\downarrow\) & mIoU\(\uparrow\) \\ \midrule
8  & 0.01 & 1.45  & 90.94 \\
16 & 0.04 & 4.83  & 91.53 \\
24 & 0.09 & 10.14 & 92.04 \\
32 & 0.16 & 17.37 & \textbf{92.45} \\
56 & 0.46 & 50.61 & 92.44 \\
64 & 0.60 & 65.54 & 92.45 \\ \bottomrule
\end{tabular}

\end{table}

\subsubsection{Ablation of PGA}
\label{sec:ab_pga}
In Tables \ref{tab:PGA_arch} and \ref{tab:PGA_channels}, we conduct ablation studies on the PGA module and analyze the parameters and FLOPs of PGA, as well as its impact on ShareCMP segmentation performance. By removing ChannelAttn, the performance of the model significantly decreases by 1.52\% in Table \ref{tab:PGA_arch}. It indicates that channel attention plays an important role in extracting polarization modal feature information from polarized images with four angles, and can effectively extract polarization modal feature information and generate polarization modal representations that are beneficial for semantic segmentation. We replace DWConv with regular convolution (Conv), and the results show that the two kinds of convolutions have similar performance in semantic segmentation. However, Conv significantly increases the FLOPs of PGA (+75.83G), reducing the computational efficiency of ShareCMP. Replacing PReLU with ReLU also reduces model performance, indicating that PReLU exhibits good performance in generating polarization modal representations, and its adaptive negative interval response is more conducive to filtering and preserving polarization modal information. Removing shortcuts from PGA results in the loss of the raw polarization modal features, thereby reducing performance. In the PGA with shallow model depth, setting a smaller dilation coefficient (dilation=1) for the first DWConv will reduce the receptive field of PGA to polarization modal features and decrease the understanding of PGA for global polarization modal features. Setting a larger dilation coefficient (dilation=3) reduces the learning of PGA for polarization modal feature continuity. Therefore, setting dilation to 2 in the PGA achieves a good balance and performance. In Table \ref{tab:PGA_channels}, we set the number of polarization image input channels \(d\) from 8 to 64, and as the number of channels increase, the FLOPs of the model significantly increase. When the number of channels is between 8 and 24, polarization modal information cannot be well expressed by low-dimensional features. When the number of channels is 32, the model trades off performance and computational complexity. As the number of channels continues to increase, the model performance also reaches saturation. Therefore, we set the polarization image input channels to 32.

\begin{table}[t]
\caption{Ablation study of the CPALoss on its settings. The \(Loss_{CPA}-\bm{A}/\bm{D}\) represents the CPALoss using AoLP/DoLP.}
\label{tab:CPALoss_settings}
\centering

\begin{tabular}{@{}cccc@{}}
\toprule
\(\lambda\) & \(Loss_{CPA}-\bm{A}\) & \(Loss_{CPA}-\bm{D}\) & mIoU\(\uparrow\) \\ \midrule
0.1           & \checkmark & \checkmark & 92.35 \\
0.05          & \checkmark & \checkmark & 92.40 \\
\textbf{0.01} & \checkmark & \checkmark & \textbf{92.45} \\
0.005         & \checkmark & \checkmark & 92.38 \\ \midrule
0.01          & \checkmark &            & 92.19 \\
0.01          &            & \checkmark & 92.21 \\ \bottomrule
\end{tabular}

\end{table}
\begin{table}[t]
\caption{Ablation study of the CPALoss on different encoder stages.}
\label{tab:CPALoss}
\centering

\begin{tabular}{@{}cccccc@{}}
\toprule
\multirow{2}{*}{ShareCMP} & \multicolumn{4}{c}{Stage} & \multirow{2}{*}{mIoU\(\uparrow\)} \\ \cmidrule(lr){2-5}
           & 1          & 2          & 3          & 4          &       \\ \midrule
w/ CPALoss & \checkmark & \checkmark & \checkmark & \checkmark & 91.79 \\
w/ CPALoss &            & \checkmark & \checkmark & \checkmark & 91.63 \\
w/ CPALoss &            &            & \checkmark & \checkmark & \textbf{92.45} \\
w/ CPALoss &            &            &            & \checkmark & 91.72 \\
w/ CPALoss &      &            &            &            & 91.09 \\ \bottomrule
\end{tabular}

\end{table}

\subsubsection{Ablation of CPALoss}
\label{sec:ab_cpaloss}
\begin{figure}[t]
\centering
\includegraphics[width=0.85\linewidth]{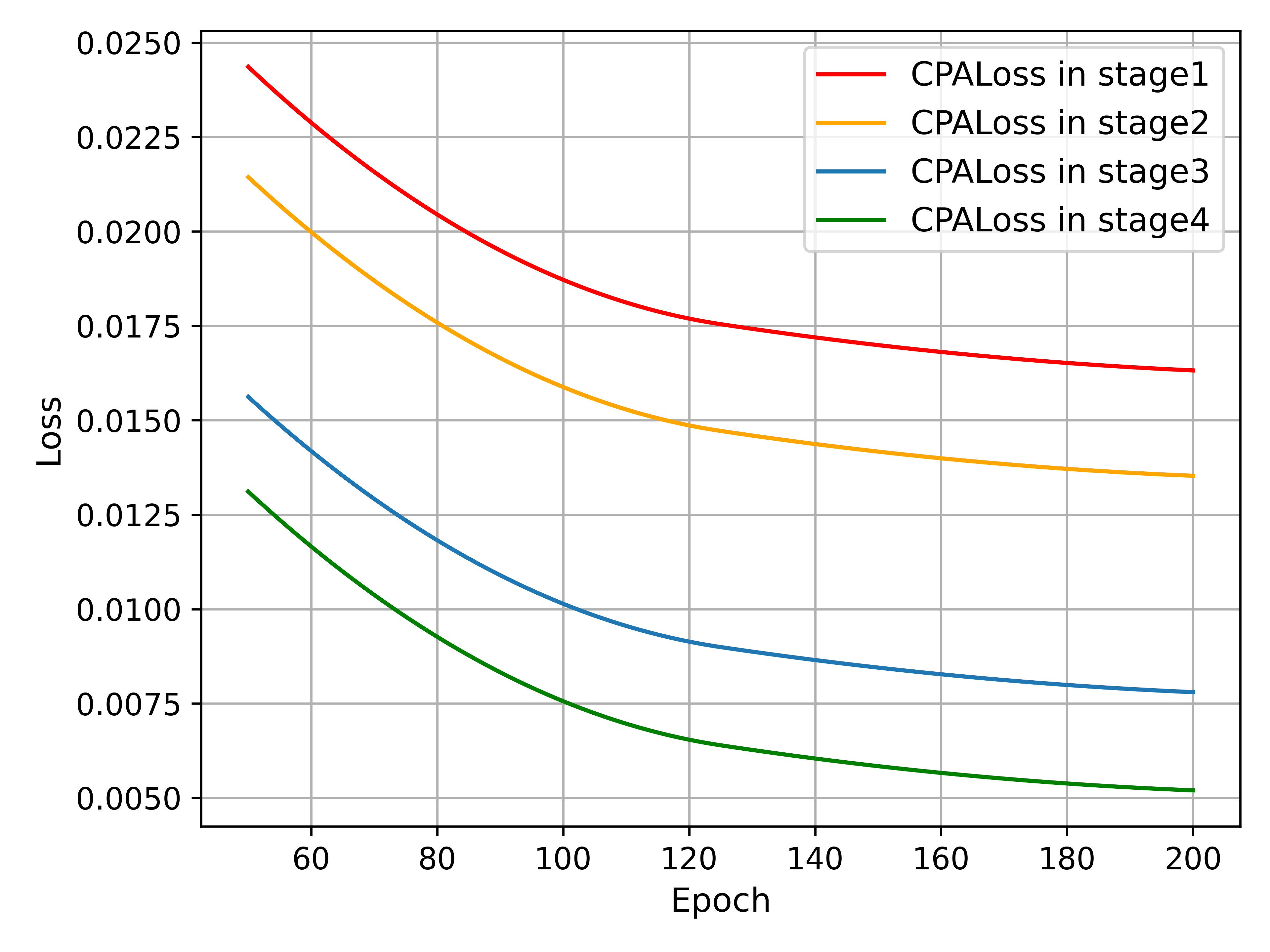}
\caption{Training convergence curves of CPALoss in different stages.}
\label{fig:CAPLoss}
\end{figure}

In Table \ref{tab:CPALoss_settings}, it can be seen that ShareCMP gets the best mIoU when \(\lambda\) is 0.01. We also ablate the CPALoss using AoLP or DoLP. The results show that the CPALoss using AoLP and DoLP performs best for ShareCMP. It provides polarization modal priori for the model, which helps the model to further understand polarization modal information. And AoLP and DoLP respectively reflect the polarization angle and polarization degree of the reflected light from the object, and both are used to optimize CPALoss. This further improves the understanding of the model for the object polarization properties. As shown in Table \ref{tab:CPALoss}, we ablate the combination of CPALoss used in different ShareCMP encoder stages and require that deep and high-level features must be optimized for CPALoss. It is difficult to extract the polarization modal information (AoLP and DoLP) of the image from the low-level features in the low-level stage, so we set CPALoss from stage 4 to 1. When the CPALoss is optimized in stages 3 and 4, ShareCMP achieves the best performance of 92.45\%. It indicates that multi-scale high-level features are beneficial for the model to learn and understand polarization properties, while CPAAHead has difficulty predicting AoLP and DoLP from low-level features. And it indicates that the features in stages 3 and 4 contain advanced information about understanding the polarization properties of the object, while the features in stages 1 and 2 have not yet constructed advanced information about understanding the polarization properties. Moreover, in Fig. \ref{fig:CAPLoss}, we visualize the training convergence curves of CPALoss in different stages. The results show that the convergence trend of CPALoss is the same in different stages, and as the stage increases, the CPALoss converges to a smaller value. But the CPALoss in stages 3 and 4 is significantly smaller than that in stages 1 and 2, which also indicates that high-level features in stages 3 and 4 are beneficial for the model to learn and understand polarization properties.

\begin{table}[t]
\caption{Ablation study of the ShareCMP framework.}
\label{tab:ShareCMP}
\centering
\resizebox{\linewidth}{!}{

\begin{tabular}{@{}lcc@{}}
\toprule
Method      & Modal     & mIoU\(\uparrow\) \\ \midrule
ShareCMP    & RGB-\{\(\bm{I}_{0}\),\(\bm{I}_{45}\),\(\bm{I}_{90}\),\(\bm{I}_{135}\)\} & \textbf{92.45} \\ \midrule
w/o ShareCMP Encoder & RGB-\{\(\bm{I}_{0}\),\(\bm{I}_{45}\),\(\bm{I}_{90}\),\(\bm{I}_{135}\)\} & 92.31{\small (-0.14)} \\
w/o CPALoss & RGB-\{\(\bm{I}_{0}\),\(\bm{I}_{45}\),\(\bm{I}_{90}\),\(\bm{I}_{135}\)\} & 91.09{\small (-1.36)} \\
w/o PGA     & RGB-AoLP  & 91.96{\small (-0.49)} \\
w/o PGA     & RGB-DoLP  & 92.03{\small (-0.42)} \\
w/o PGA     & RGB-SAoLP & 90.63{\small (-1.82)} \\
w/o PGA     & RGB-CAoLP & 91.15{\small (-1.30)} \\ \bottomrule
\end{tabular}

}
\end{table}
\begin{figure*}[t]
\centering
\includegraphics[width=1.0\linewidth]{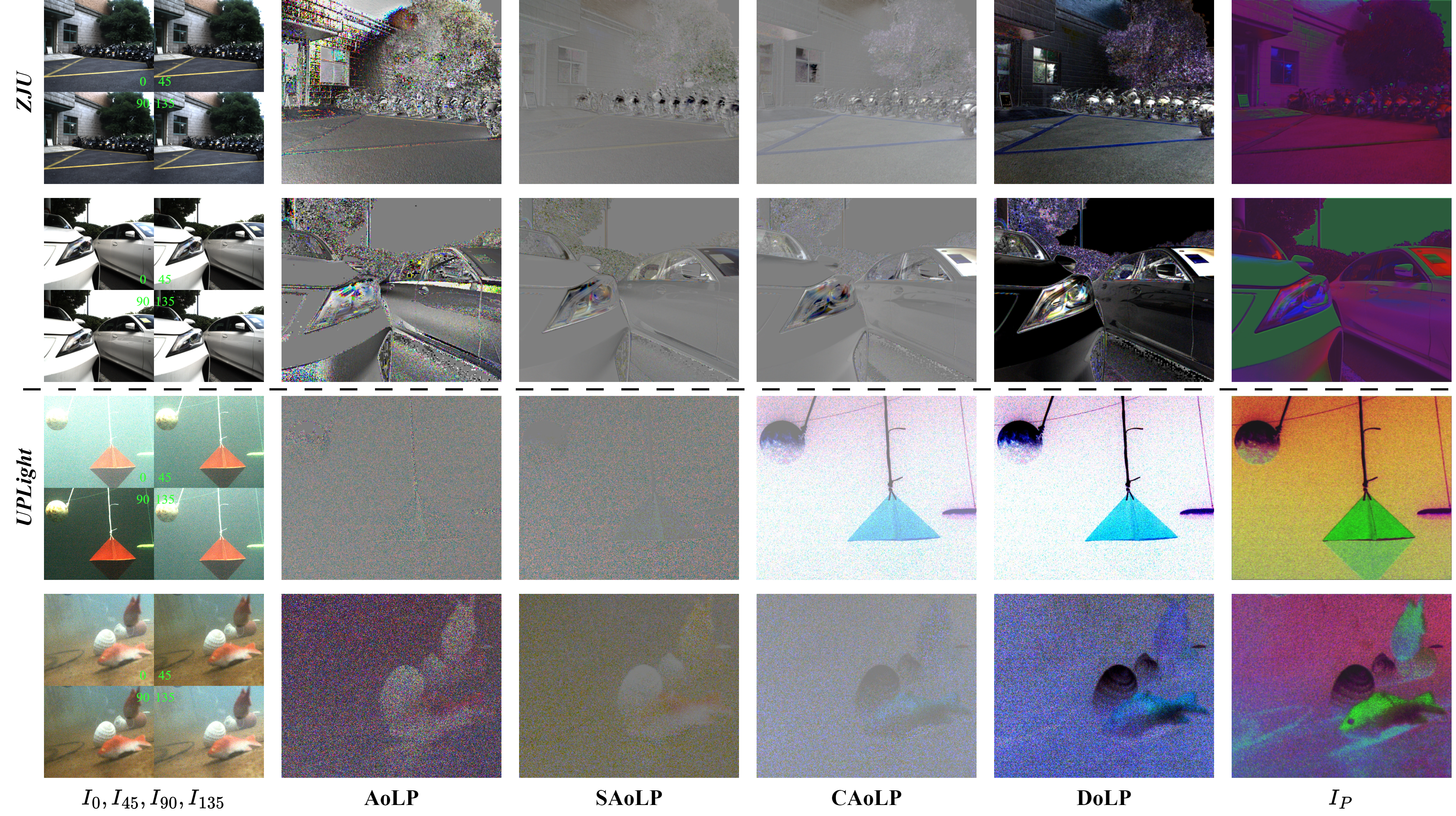}
\caption{Visualization of different polarization modal representations SAoLP and CAoLP, AoLP, DoLP and \(\bm{I}_{P}\) which is generated by the PGA module. Compared with other polarization modal representations, \(\bm{I}_{P}\) has richer polarization features, and different polarization features display different colors.}
\label{fig:p_modal}
\end{figure*}

\subsubsection{Ablation of ShareCMP}
In Table \ref{tab:ShareCMP}, we show the semantic segmentation performance of ShareCMP without ShareCMP Encoder, CPALoss, or PGA and investigate the effect of the four polarization modalities AoLP, DoLP, SAoLP, and CAoLP on the RGB-P semantic segmentation performance. The performance decreases by 0.14\% when without the shared parameters, which indicates that the ShareCMP encoder not only reduces the model parameters, but also improves the performance of the model. The ShareCMP encoder extracts features from two modalities (RGB-P), achieving consistency in feature distribution and semantic representation between the two modalities, which facilitates feature fusion between them. When removing the CPALoss for polarization-aware learning, the performance decreases significantly by 1.36\%. Our designed SAoLP and CAoLP, which are new representation methods for the polarization modality, result in poor performance decreases of 1.82\% and 1.30\%, respectively. Since SAoLP and CAoLP are part of AoLP, their polarization modal representation is not as complete as AoLP, resulting in reduced performance. While AoLP and DoLP result in a similar slight decrease in performance, this indicates that AoLP and DoLP have similar effects on the ShareCMP and are also relatively effective polarization modal representations. Regardless of the fixed paradigm of the polarization modal representation, its performance is not as good as the representation generated by our PGA. The results confirm that our ShareCMP Encoder, CPALoss, and PGA are more advantageous for RGB-P semantic segmentation.

\begin{table}[t]
\caption{Comparison of the number of parameters (\#Params), FLOPs, FPS, and Memory which are counted in \(512\times 512\). The models are deployed on NVIDIA RTX A6000 by ONNX (GPU) FP32 and NVIDIA Jetson Orin Nano by ONNX (GPU) FP16. ShareCMP* represents the ShareCMP w/o PGA module and CPALoss.}
\label{tab:deploy}
\centering
\resizebox{\linewidth}{!}{

\begin{tabular}{@{}lcccccc@{}}
\toprule
Method & Device & \#Params (M)\(\downarrow\) & FLOPs (G)\(\downarrow\) & Memory (MB)\(\downarrow\) & FPS\(\uparrow\) \\ \midrule
CMNeXt~\cite{cmnext} & A6000     & 58.73 & 65.42 & 224.02 & 17.4 \\
CMX~\cite{cmx}       & A6000     & 65.52 & 52.03 & 249.92 & 25.2 \\
ShareCMP             & A6000     & 43.40 & 66.19 & 167.21 & 17.2 \\
ShareCMP*            & A6000     & \textbf{43.24} & \textbf{51.66} & \textbf{166.59} & \textbf{25.3} \\ \midrule
CMNeXt               & Orin Nano & 58.73 & 65.42 & 112.01 & 6.36 \\
CMX                  & Orin Nano & 65.52 & 52.03 & 124.96 & 8.16 \\
ShareCMP             & Orin Nano & 43.40 & 66.19 & 83.61  & 6.13 \\
ShareCMP*            & Orin Nano & \textbf{43.24} & \textbf{51.66} & \textbf{83.30}  & \textbf{8.25} \\ \bottomrule
\end{tabular}

}
\end{table}

\subsection{Quantitative and Visual Analysis}
\label{sec:qv_analysis}

\subsubsection{Quantitative analysis of deployed ShareCMP}
In Table \ref{tab:deploy}, we compare models deployed on A6000 and Orin Nano devices by ONNX (GPU). Our ShareCMP reduces the number of parameters by about 22 M compared to CMX~\cite{cmx}. The model parameters of ShareCMP only have 166.59 MB (on A6000) and 83.30 MB (on Orin Nano) of memory space, which reduces the parameters and memory space by 33.8\% compared to CMX. And our ShareCMP* (without PGA module and CPALoss) achieves 25.3 (on A6000) and 8.25 (on Orin Nano) FPS. The test of ShareCMP* demonstrates its performance when extended for use on RGB-X data (not RGB-P). Although the FPS of ShareCMP has not been significantly improved, its parameter count and memory usage have been significantly reduced, which is friendly to embedded devices with small running memory. During the model inference, it occupies too much running memory, which can cause the running memory to overflow and interrupt. In practical applications, the perception module of AUVs is deployed with multiple processing models for multiple tasks, so the model occupies less running memory, which is an excellent advantage for model deployment in AUVs. Our main contribution is to improve the memory usage of the model during deployment and runtime.

\subsubsection{Visualization of polarization modal representations}
We show five polarization modal representations, as in Fig. \ref{fig:p_modal}. Compared with SAoLP and CAoLP, AoLP and DoLP cover the polarization information in SAoLP and CAoLP and show richer polarization features, which also verifies that the segmentation results of RGB-AoLP and RGB-DoLP are better than RGB-SAoLP and RGB-CAoLP in Table \ref{tab:ShareCMP}. While \(\bm{I}_{P}\) has rich polarization properties, it also provides different color features for objects with different polarization properties, such as the glass reflection and the reflector of the motorcycle, the car shell at different angles in the second group of images, the glass and the sticker on the car glass on the right side, and different colors of classes in the third and fourth groups of images. The AoLP and SAoLP in the third group of images only have very few polarization features, while our \(\bm{I}_{P}\) can still provide rich polarization features, indicating that our PGA module has good robustness in generating polarization features. In addition, compared to \(\bm{I}_{P}\), the other four polarization modal representations are very coarse in some areas and have a lot of noise points, while \(\bm{I}_{P}\) has smoother polarization features that eliminate a large number of noise points, especially for the underwater polarization modal representations in the UPLight dataset.

\begin{figure}[t]
\centering
\includegraphics[width=1.0\linewidth]{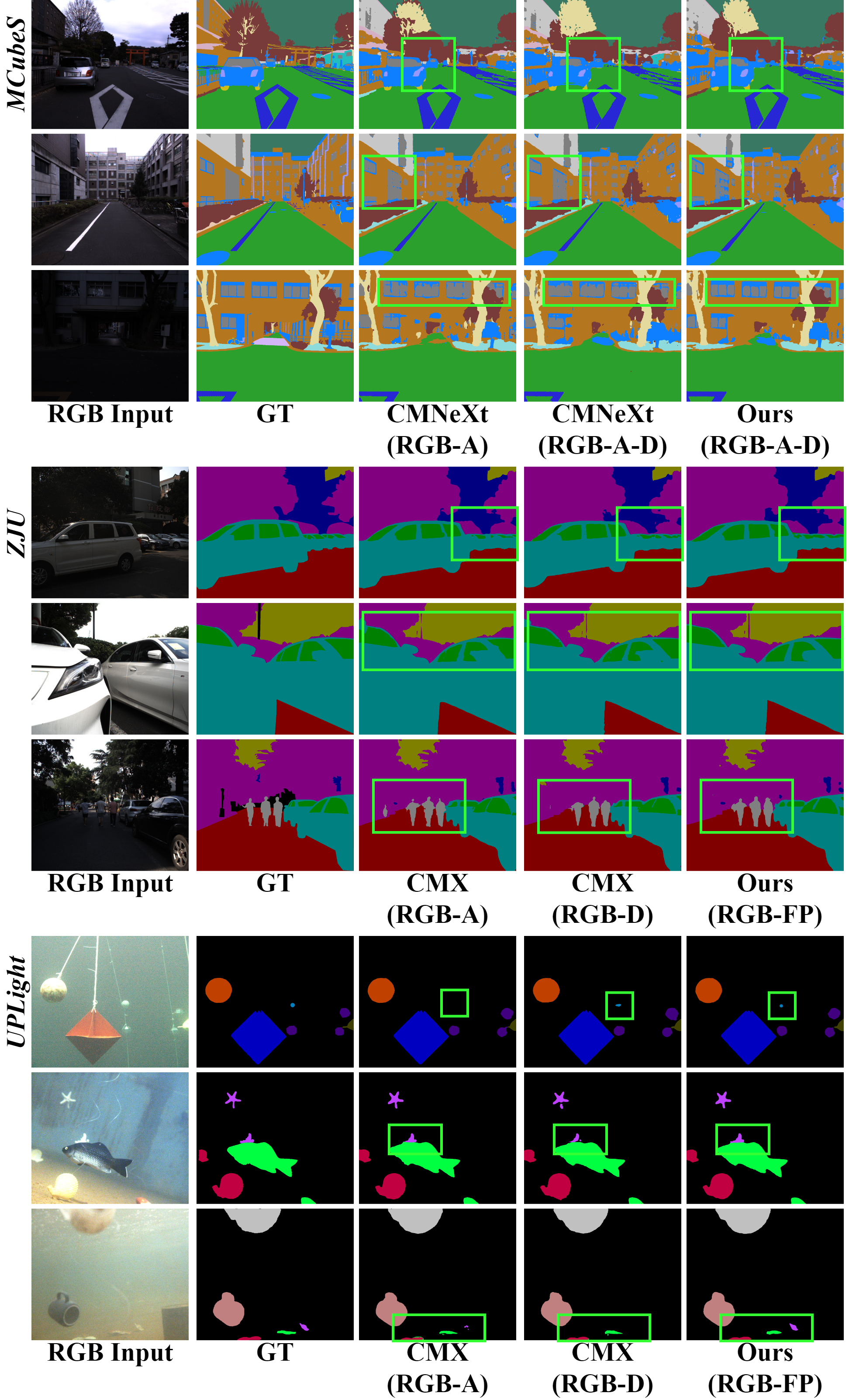}
\caption{Visualization of segmentation results. RGB-FP represents RGB and four polarized images with different polarization angles (RGB-\{\(\bm{I}_{0}\),\(\bm{I}_{45}\),\(\bm{I}_{90}\),\(\bm{I}_{135}\)\}).}
\label{fig:seg_vis}
\end{figure}

\subsubsection{Visualization of segmentation results}
In Fig. \ref{fig:seg_vis}, we show the semantic segmentation results of our ShareCMP against the CMX~\cite{cmx} with RGB-A and RGB-D and the CMNeXt~\cite{cmnext} with RGB-A and RGB-A-D. It can be seen that our ShareCMP achieves a more complete segmentation for glass windows, car light on MCubeS~\cite{mcubes} and glass windows, persons on ZJU~\cite{zju} images, in the case of insufficient illumination. For blurred underwater images in UPLight, CMX does not segment the starfish in the image or segments the starfish, shells, and floats incompletely. Our ShareCMP is more robust in underwater scenes and accurately segments objects, and ShareCMP can better understand the polarization properties of object materials, such as glass and calcium materials.
\section{Conclusion}
\label{sec:conclusion}

In this work, we propose ShareCMP for RGB-P multimodal underwater semantic segmentation. Our ShareCMP, a shared dual-branch architecture with ShareCMP Encoder, reduces the parameters and memory space by about 33.8\% compared to CMX. The experimental results indicate that the PGA module can generate polarization modal images with richer polarization properties compared to AoLP and DoLP. In addition, the CPALoss improves the learning and understanding of the encoder for polarization modal information. Our ShareCMP achieves the best performance with fewer parameters on the three RGB-P datasets.

The shared dual-branch architecture can also be applied to other multimodal fields (RGB-X) to reduce the number of model parameters without sacrificing performance. However, our ShareCMP Encoder reduces the number of parameters but not the computational cost, which is limited by the fact that the two modal features must be extracted to the same level for cross-attention fusion. In future work, we aim to investigate a novel single-branch multimodal backbone as an alternative to overcome the limitations of parameter and computational complexity associated with the dual-branch backbone.
\medskip
\newline
\noindent
\textbf{Acknowledgments.}
This research is funded by the National Natural Science Foundation of China, grant number 52371350, by the National Key Research and Development Program of China, grant number 2023YFC2809104, and by the National Key Laboratory Foundation of Autonomous Marine Vehicle Technology, grant number 2024-HYHXQ-WDZC03.
{
    \small
    \bibliographystyle{ieeenat_fullname}
    \bibliography{sections/main}
}


\end{document}